\documentclass[twoside]{article}

\usepackage[accepted]{aistats2024}
\usepackage[round]{natbib}
\renewcommand{\bibname}{References}
\renewcommand{\bibsection}{\subsubsection*{\bibname}}

\usepackage{amsmath,amsfonts,bm}

\def\eqref#1{equation~\ref{#1}}

\def\1{\bm{1}}

\DeclareMathAlphabet{\mathsfit}{\encodingdefault}{\sfdefault}{m}{sl}
\SetMathAlphabet{\mathsfit}{bold}{\encodingdefault}{\sfdefault}{bx}{n}

\let\ab\allowbreak

\usepackage{hyperref}
\usepackage{url}

\usepackage{tabularray}
\UseTblrLibrary{booktabs}
\usepackage{derivative}
\usepackage{nicematrix}
\usepackage{array}
\usepackage{microtype}
\usepackage{graphicx}
\usepackage{subfigure}
\usepackage{caption}
\usepackage{subcaption} 
\usepackage{booktabs} %
\definecolor{gray}{rgb}{.6,.6,.6}
\renewcommand{\paragraph}[1]{\noindent\textbf{#1}\quad}

\renewcommand{\ab}[1]{{\color{blue}{{\bf Ahmad}: #1}}}
\newcommand{\edits}[1]{{\color{black}{#1}}}

\newcounter{remark}[section]
\newenvironment{remark}[1][]{\refstepcounter{remark}\par
\noindent \textbf{Remark~\theremark.#1} \rmfamily}{}

\definecolor{ruddy}{rgb}{1.0, 0.0, 0.16}
\definecolor{gblue}{RGB}{29, 144, 255}
\definecolor{royalblue}{rgb}{0.25, 0.41, 0.88}
\definecolor{edits}{rgb}{1.0,0.0,0.0}

\hypersetup{
  colorlinks,
  citecolor=royalblue,
  linkcolor=ruddy,
  urlcolor=royalblue}

\usepackage{comment}

\newcommand{\ba}{{{\bf a}}}

\newcommand{\bC}{{{\bf C}}}

\newcommand{\be}{{{\bf e}}}

\newcommand{\bI}{{{\bf I}}}
\renewcommand{\bm}{{{\bf m}}}

\newcommand{\bn}{{{\bf n}}}
\newcommand{\bN}{{{\bf N}}}

\newcommand{\bs}{{{\bf s}}}

\newcommand{\bu}{{{\bf u}}}

\newcommand{\bw}{{{\bf w}}}

\newcommand{\bx}{{{\bf x}}}

\newcommand{\by}{{{\bf y}}}

\newcommand{\bzero}{{\boldsymbol{0}}}

\newcommand{\EQ}{\begin{equation}}
\newcommand{\EN}{\end{equation}}
\newcommand{\ben}{\begin{enumerate}}
\newcommand{\een}{\end{enumerate}}

    \def\squarebox#1{\hbox to #1{\hfill\vbox to #1{\vfill}}}

\begin{document}

\runningtitle{Improving Robustness via Tilted Exponential Layer}

\runningauthor{Bhagyashree Puranik, Ahmad Beirami, Yao Qin, Upamanyu Madhow}

\twocolumn[

\aistatstitle{
{Improving Robustness via Tilted Exponential Layer:\\ A Communication-Theoretic Perspective}}

\aistatsauthor{Bhagyashree Puranik$^{1}$ \And Ahmad Beirami$^{2}$ \And  Yao Qin$^{1,2}$ \And Upamanyu Madhow$^{1}$ }

\aistatsaddress{$^{1}$University of California Santa Barbara \And  $^{2}$Google Research } ]

\begin{abstract}
\vspace{-0.2in}
State-of-the-art techniques for enhancing robustness of deep networks mostly rely on empirical risk minimization with suitable data augmentation. In this paper, we propose a complementary approach motivated by communication theory, aimed at enhancing the {\em signal-to-noise ratio} at the output of a neural network layer via neural competition during learning and inference. In addition to standard empirical risk minimization, neurons compete to sparsely represent layer inputs by maximization of a tilted exponential (TEXP) objective function for the layer. TEXP learning can be interpreted as maximum likelihood estimation of {\em matched filters} under a Gaussian model for {\em data noise}. Inference in a TEXP layer is accomplished by replacing batch norm by a tilted softmax, which can be interpreted as computation of posterior probabilities for the competing signaling hypotheses represented by each neuron. After providing insights via simplified models, we show, by experimentation on standard image datasets, that TEXP learning and inference enhances robustness against noise and other common corruptions, without requiring data augmentation. Further cumulative gains in robustness against this array of distortions can be obtained by appropriately combining TEXP with data augmentation techniques. %
The code for all our experiments is available at~\url{https://github.com/bhagyapuranik/texp_for_robustness}.

\end{abstract}

\section{Introduction}
\vspace{-.14in}
Standard training of deep neural networks is well known to lack robustness against a variety of distortions, including noise, distribution shifts \citep{hendrycks2018benchmarking, dodge2017study}, and adversarial attacks \citep{szegedy2013intriguing,goodfellow2014adversarial,carlini2016distillationrefuted}. The most common approach to improving robustness relies on performing data augmentation. For example, adversarial training \citep{madry2017towards}, which augments the training data with generated adversarial examples (corresponding to the current realization of the network parameters), is one of the most effective adversarial defenses against adversarial attacks. In addition, different types of data augmentation have also been shown to effectively improve robustness against natural corruptions~\citep{cubuk2019autoaugment,hendrycks2020Augmix, qin2023effective}.

In this paper, in a manner that is {\it complementary} to learning with data augmentation, we propose and explore a strategy for enhancing robustness based on detection and estimation theoretic concepts, motivated by their success in fields such as wireless communication systems.  In communication theory, the receiver tries to match the
incoming signal against a number of possible signal templates, each corresponding to a different message. For signaling in Gaussian noise, correlating against these signal templates, often called matched filtering, maximizes the signal-to-noise ratio, and the posterior probability of each possible transmitted signal is obtained by feeding suitably scaled matched filter outputs to a softmax.  We propose to apply these ideas to enhance the {\em signal-to-noise ratio} in a neural network layer (our exploration here focuses on the first layer) via neuronal competition during learning and inference.

Unlike in communication systems, we do not have a known set of messages and corresponding transmitted symbols.  Rather, we seek to learn layer weights which are well matched to the set of incoming patterns, so that for each strong input, a fraction of neurons fire strongly.  We accomplish this by adding tilted exponential objective for the layer, which can be interpreted as maximum likelihood estimation of {\em matched filter} signal templates under a Gaussian model for ``data noise''  (see Sec.~\ref{sec:comm_theory_motivation}). For inference, we replace batch norm by a tilted softmax, interpretable as computation of posterior probabilities for competing signal templates represented by the neurons. Our framework allows us to vary the amount of {\em data noise} during training (smaller if training with clean data) and during inference (bigger to provide robustness against out of distribution ``noise'').  We term a layer designed in this way (see Sec.~\ref{sec:texp_dnn}) as a tilted exponential (TEXP) layer.  

We provide geometric insight into TEXP training for simplified models, and then demonstrate enhancements in robustness for CNNs operating on standard image datasets.
Experiments on CIFAR-10~\citep{krizhevsky2009learning} show that replacing the first layer of a VGG-16~\citep{simonyan2014very} network by a TEXP layer yields increased robustness against noise, other common corruptions and mild adversarial perturbations {\it without requiring data augmentation}. Additional performance gains are obtained by supplementing TEXP with adversarial training and other data augmentation techniques such as AugMix~\citep{hendrycks2020Augmix}, RandAugment~\citep{cubuk2020randaugment} and AutoAugment~\citep{cubuk2019autoaugment}. We show that the TEXP approach generalizes to other network architectures and datasets through promising preliminary experiments on CIFAR-100 (using Wide-ResNet-28-10 as backbone) and ImageNet (ResNet-50 backbone).

\vspace{-.1in}
\section{Related work} \label{sec:related}
\vspace{-.1in}
Disparity between the data observed during training and testing phases is a common phenomenon, highlighting the significance of robustness in generalizing to out-of-distribution (OOD) samples. The prevailing approach to address this challenge is to use various forms of OOD data augmentation ~\citep{zhang2018mixup,cubuk2019autoaugment,cubuk2020randaugment,schneider2020improving,hendrycks2020Augmix,calian2022defending,kireev2022effectiveness,qin2022understanding}, often combined with techniques such as proxy tasks or consistency regularization.
For example, Augmix~\citep{hendrycks2020Augmix} enriches training images by incorporating a composition of randomly sampled augmentations to generate a diverse set of augmented images, supplemented by a consistency loss function aimed at encouraging DNN outputs to react smoothly to augmentations.
Such consistency regularization for data augmentation has shown promise in several other works as well~\citep{tack2022consistency,huang2022robustness}.

A complementary set of works~\citep{gilmer2019adversarial, yi2021improved, yin2019fourier, qin2023effective} indicate that adversarial training (typically with small perturbation budgets) can also lead to improvements in OOD robustness. However, finding techniques that work well for various kinds of OOD corruptions, particularly without heavy data augmentation, remains challenging. For example,~\cite{yin2019fourier} find that adversarial training and Gaussian noise augmentation improve robustness against certain corruptions like other types of noise and blurs while degrading the performance under \textit{low frequency} corruptions like fog and contrast. They argue that a diverse set of augmentations may be required to combat such trade-offs. Our TEXP method shows promise in achieving broad spectrum robustness without data augmentations, while also combining well with strategies such as AugMix and adversarial training.

Our approach of adding a cost based on layer activations is motivated by the recent work~\citep{cekic}, which argues that targeting sparse, strong activations at early network layers can improve robustness. \citet{cekic} employs Hebbian/anti-Hebbian (HaH) training at the layers, in which the most active neurons for an input are promoted towards the input (``fire together, wire together''), while neurons which are less active are demoted away from the input, and uses divisive normalization (enabling smaller outputs to be attenuated by larger outputs) instead of batch norm for inference. In contrast to neuroscientific motivation in HaH, our TEXP training and inference approach is derived from communication-theoretic foundations. Our approach also promotes neuronal competition in both training and inference, and results in sparse, strong activations. However, our framework leads to smoother objective functions, does not require sorting in either training or inference, and our best schemes substantially outperform~\citet{cekic}.

Exponential tilting is well-known in statistics for rejection sampling, rare-event simulation, saddle-point approximation~\citep{butler2007saddlepoint}, and importance sampling~\citep{siegmund1976importance}. It is also at the heart of Chernoff bounds~\citep{dembo-zeitouni}, as well as analyzing atypical events in information theory~\citep{beirami2018characterization}, and has appeared as a smoothing method to maximum in optimization literature~\citep{kort1972new,pee2011solving,liu2019deep}.
These concepts have motivated prior work on tilted exponentials applied to the {\em training objective function,} which has been demonstrated to yield 
fairness and robustness benefits in a multitude of machine learning problems~\citep{li2021tilted,li2023tilted}.  
Unlike this prior work on exponential tilting, which is motivated by connections to Chernoff bounds, large deviations and typicality, our
proposal of TEXP objective is motivated by maximum likelihood estimation of signal templates, and we apply exponential tilting to layer activations.

\vspace{-.1in}
\section{\hspace{-0.1cm}{Learning matched filters with TEXP}}
\label{sec:comm_theory_motivation}
\vspace{-.1in}
We provide here a communication-theoretic motivation for training and inference in a TEXP layer, and provide insight into why it produces sparse, strong activations, which in turn provide robustness. We also provide a geometric illustration of the neuronal weights obtained with TEXP training via simplified models.

A classical model in communication theory is to consider the received signal as one of $M$ possible transmitted messages, corrupted by white Gaussian noise. Under 
hypothesis $H_i$, the received signal is modeled as
\vspace{-0.3cm}
\EQ \label{hypotheses}
H_i: \bx = \bw_i + \bn 
\EN
where $\{\bw_i\}_{i \in [M]}$ (${[M] := \{1, \ldots, M\}}$) are the $M$ possible known messages, and $\bn$ is white Gaussian noise with variance $\nu^2$ per dimension.

In this setting, it is well known~\citep{madhow2008fundamentals} that sufficient statistics for optimal reception are obtained by correlating the received signal against each of the possible transmitted signals. That is, the optimal receiver computes the inner products $\bx^T \bw_i$, $i \in [M]$, to make its decisions. These correlation operations are often termed {\em matched filtering,} since they try to match the received signal against one of $M$ possible templates.  

The proposed tilted exponential (TEXP) approach for robustness is based on fitting the model (\ref{hypotheses}) to the input $\bx$ to a layer in a neural network (the experimental results presented in this paper focus exclusively on the first layer of a convolutional neural network, but in principle, the concepts could be applied to any layer of a neural network).  
For a given layer of a neural network with $M$ neurons (or $M$ output channels or filters in case of a CNN), we interpret the neuronal weights $\mathcal{W}={\{\bw_i\} }_{i \in [M]}$ as signal templates corresponding to $M$ possible hypotheses regarding the input $\bx$, with the $i$th neuron producing the matched filter output $\bx^T \bw_i$, $i \in [M]$.  However, unlike in a communication system, we do not know the set of possible ``messages''  \emph{a priori.}  We must therefore learn the ``matched filters'' $\mathcal{W}$ based on input training samples during training, and then perform inference based on these learnt templates.  Naturally, we do not expect such a model to be accurate, but fitting it to data provides an approach for learning neural weights such that, for each input, it is likely that there is a subset of neurons well matched to it.  The parameter $\nu^2$ may be viewed as ``data noise,'' acknowledging that the input $\bx$ may not fit any of the templates we learn.

\paragraph{\bf Likelihood function.} For the model (\ref{hypotheses}), the conditional density of $\bx$ under hypothesis $H_i$ is $p(\bx|H_i) = N(\bx| \bw_i , \nu^2 \bI)$, where $N(\bx| {\boldsymbol{\mu}} , \boldsymbol{\Sigma})$ is the Gaussian density with mean ${\boldsymbol{\mu}}$ and covariance $\boldsymbol{\Sigma}$. However, a more convenient representation for maximum likelihood estimation is to take its Radon-Nikodym derivative with respect to the conditional density $N(\bx| \mathbf{0} , \nu^2 \bI)$ for a ``noise-only'' dummy hypothesis, to obtain the conditional likelihoods:
\vspace{-0.2cm}
\begin{eqnarray}
L_{\mathcal{W}} ( \bx |H_i ) &=& \frac{N(\bx| \bw_i , \nu^2 \bI)}{N(\bx| \mathbf{0} , \nu^2 \bI)} \nonumber
\end{eqnarray}
\begin{eqnarray} \vspace{-0.3cm}
\hspace{1cm}&=&  \exp \left( \frac{1}{\nu^2} ( \bx^T \bw_i - \|\bw_i\|^2/2 ) \right),  \label{conditional_likelihoods} \vspace{-0.2cm}
\end{eqnarray}
for $i \in [M]$, where we now view $\mathcal{W}$ as parameters to be learnt during training.  We see that these conditional likelihoods depend on the input $\bx$ only through the matched filter outputs $\bx^T \bw_i$, $i \in [M]$, which means that these are sufficient statistics.

Assuming that all templates have equal energy, we can drop the $ \|\bw_i\|^2/2 $ terms from (\ref{conditional_likelihoods}) to obtain the simplified expression, for all $i \in [M]:$
\vspace{-0.2cm}
\EQ \label{conditional_likelihoods2}
L_{\mathcal{W}} ( \bx |H_i ) = \exp \left( \frac{1}{\nu^2}  \bx^T \bw_i \right).
\EN
Averaging over the conditional likelihoods (\ref{conditional_likelihoods2}), the likelihood of $\bx$ is obtained as a sum of tilted exponentials:
\vspace{-0.2cm}
\EQ \label{likelihood}
L_{\mathcal{W}} ( \bx) = \frac{1}{M} \sum_{i=1}^M \exp \left( \frac{1}{\nu^2}  \bx^T \bw_i  \right) = \frac{1}{M} \sum_{i=1}^M \exp \left( t a_i  \right),
\EN
where $t = \frac{1}{\nu^2} > 0$ is the tilt parameter and $a_i =  \bx^T \bw_i $ is the activation (or matched filter output) produced by the $i$-th neuron.

\paragraph{\bf TEXP training objective.} Maximum likelihood estimation of signal templates using a collection of independently drawn data points is accomplished
by maximizing the sum of log-likelihoods. This corresponds to the following tilted exponential objective function:  \vspace{-0.4cm}
\EQ \label{log_likelihood}
\mathcal{L}_{\text{TEXP}} ( \bx ) = \log L_{\mathcal{W}} ( \bx) =  \log \frac{1}{M} \sum_{i=1}^M \exp \left( t a_i  \right).\vspace{-0.2cm}
\EN

\paragraph{\bf Implicit normalization of templates.} While different transmitted signals in a communication system might have different energies, we wish to enforce fair competition
across templates by normalizing each to unit norm.  This can be accomplished without explicit optimization constraints via
implicit normalization of activations, by redefining as: $a_i = \frac{\bx^T \bw_i}{\|\bw_i\|_2} $, $i \in [M]$.  

\paragraph{\bf Notation (softmax).} Before computing the gradient of the TEXP objective, we recall the standard notation $\sigma (\bu ) = \left( \sigma_1 (\bu),...,\sigma_M (\bu) \right)^T$ for a softmax operating on a vector $\bu = (u_1,...,u_M)^T$, with
\EQ \label{softmax_defn}\vspace{-0.3cm}
\sigma_i (\bu ) = \frac{e^{u_i}}{\sum_{j=1}^M e^{u_j}} ~, ~ i=1,...,M 
\EN

\paragraph{\bf TEXP gradient.} The gradient of the objective function (\ref{log_likelihood}) is obtained as\vspace{-0.1cm} \small
\EQ \label{texp_gradient} 
\nabla_{\mathcal{W}} \mathcal{L}_{\text{TEXP}} = t\sum_{i=1}^M \frac{e^{ta_i}}{\sum_{j=1}^M e^{ta_j}} \nabla_{\mathcal{W}} a_i = t\sum_{i=1}^M \sigma_i (t \ba) \nabla_{\mathcal{W}} a_i,
\EN \normalsize
Since larger activations are weighted more via the tilted softmax, gradient ascent corresponds to increasing larger activations further.

\begin{figure}
    \centering \vspace{-0.3cm}
    \includegraphics[width=\columnwidth]{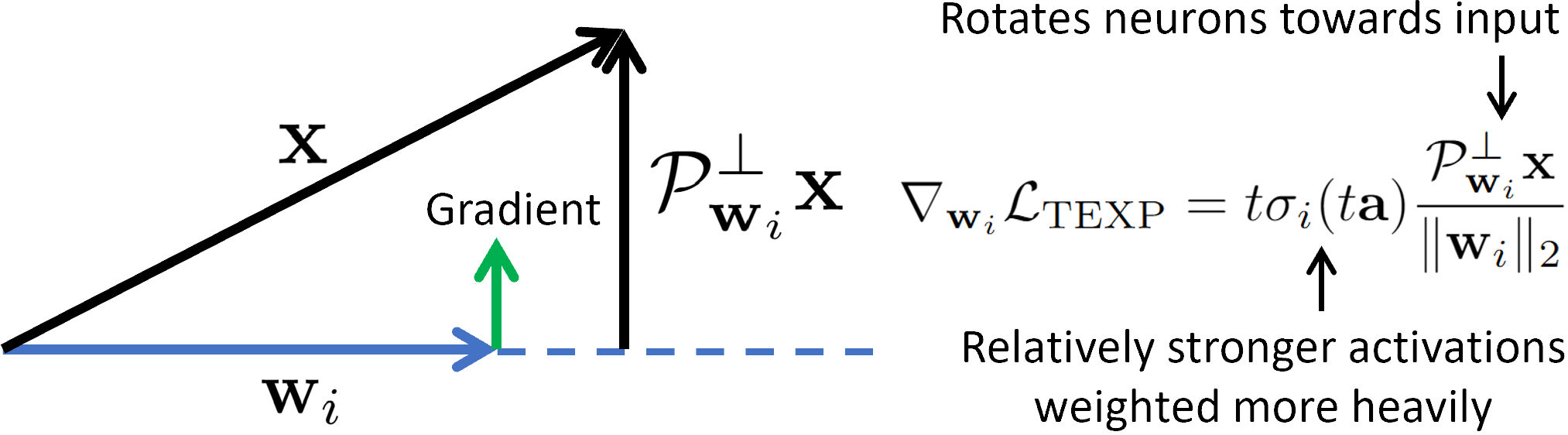}    
    \caption{TEXP gradient ascent rotates the neuron towards the input, with larger activations weighted more via a tilted softmax.}
    \label{fig:texp_update_geometry}\vspace{-0.5cm}
\end{figure}

Accounting for implicit normalization, we obtain the gradient 
\vspace{-0.2cm} 
\EQ \label{texp_gradient2}
\nabla_{\bw_i} a_i = \nabla_{\bw_i}  \frac{\bx^T \bw_i}{\|\bw_i\|_2} = \frac{{\cal{P}}^{\perp}_{{\bw}_i} \bx}{\|\bw_i\|_2}, \nonumber \text{  where,}
\EN
\vspace{-0.1cm} 
$$
{\cal{P}}^{\perp}_{{\bw}_i} \bx = \bx -  \left( \frac{\bx^T \bw_i}{\|\bw_i\|_2} \right) \frac{\bw_i}{\|\bw_i\|_2}
$$
is the projection of the input $\bx$ orthogonal to the one-dimensional subspace spanned by $\bw_i$ (Fig.~\ref{fig:texp_update_geometry}).
Note that, for $j \neq i$, $\nabla_{\bw_j} a_i = 0$.

We can now write $\nabla_{\mathcal{W}} \mathcal{L}_{\text{TEXP}} = (\nabla_{\bw_1} \mathcal{L}_{\text{TEXP}},..., $ $\nabla_{\bw_M} \mathcal{L}_{\text{TEXP}} )$, where
\vspace{-0.3cm} 
\EQ \label{texp_gradient3}
\nabla_{\bw_i} \mathcal{L}_{\text{TEXP}} = t \sigma_i(t \ba) \frac{{\cal{P}}^{\perp}_{\bw_i} \bx}{\|\bw_i\|_2}.
\EN
We now see in explicit form that a TEXP gradient update rotates each neuron to align more closely with the input (see Fig.~\ref{fig:texp_update_geometry}), with the softmax weighting favoring templates that are better aligned, so that large activations are made even larger.

\paragraph{\bf Balanced TEXP objective function.}  Additional competition among the signal templates seeking to fit an input can be created by imposing a \textit{balance} constraint in which the mean of the signal templates is set to zero.  That is, we replace $\bw_i$ by $\bw_i - \Bar{\bw}$, for $i\in [M]$, where $\Bar{\bw} = (1/M)\sum_{i=1}^M \bw_i$. This yields a
variant of (\ref{log_likelihood}) which we term a \emph{balanced} tilted exponential objective function:
\vspace{-0.4cm} 
$$
\mathcal{L}_{\text{TEXP}}^{\text{bal}} ( \bx ) =  \log \frac{1}{M} \sum_{i=1}^M \exp \left( t (a_i - \Bar{a})  \right), \vspace{-0.2cm} 
$$
where $\Bar{a} = (1/M) \sum_{i=1}^M a_i$ is the mean activation of all neurons. The corresponding gradient components are given by
\vspace{-0.4cm} 
\EQ \label{texp_gradient_bal}
\nabla_{\bw_i} \mathcal{L}_{\text{TEXP}}^{\text{bal}} = t \left( \sigma_i(t\ba) - 1/M \right) \frac{{\cal{P}}^{\perp}_{\bw_i} \bx}{\|\bw_i\|_2}
\EN
Now, in addition to making large activations larger by rotating towards the input, we make small activations smaller (i.e., such that tilted softmax is smaller than $1/M$) by 
by rotating the corresponding template \emph{away} from the input (for geometric insights, see Sec~\ref{sec:geom_insights}). %

\paragraph{\bf TEXP inference.} Once we learn the estimates of the signal templates $\mathcal{W}$, inference based on a data point $\bx$ consists of computing the posterior probability of each hypothesis (this is termed ``soft decisions'' in communication systems). For hypothesis $H_i$, this posterior probability is given by the softmax:
\vspace{-0.2cm} 
\EQ \label{TEXP_inference}
\small
p_i ( \bx )= \frac{L_{\mathcal{W}} ( \bx |H_i ) P(H_i)}{\sum_{j=1}^M L_{\mathcal{W}} ( \bx |H_j ) P(H_j)} = \sigma_i \left( \ba/\nu^2 \right) = \sigma_i (t \ba)
\EN
setting $t= \frac{1}{\nu^2}$.

\paragraph{\bf Different tilt parameters for training and inference.}
The value of $\nu^2$ used during inference using (\ref{TEXP_inference}) may be different from that for training as in (\ref{log_likelihood}).  In particular, we may
use a smaller value of $\nu^2$ (higher $t$) during training, where we might be learning from clean data, or from data that we have perturbed in a controlled manner.  On the other hand, during inference, we may use a higher value of $\nu^2$ (lower $t$) in order to accommodate data noise due to a variety of distortions that were not present during training. 
Note that TEXP inference (\ref{TEXP_inference}) is unaffected by whether or not the signal templates are balanced, since balancing corresponds to subtracting the same constant from each activation.

\vspace{-0.2cm}
\subsection{Why TEXP is expected to reduce sensitivity to perturbations.} \vspace{-0.2cm}
TEXP training pushes activations from different neurons apart, nudging different neurons to align with different signal templates.  This combines well with TEXP inference: the softmax nonlinearity enables large activations, corresponding 
to neurons well-aligned with the input, to suppress smaller activations, and reduce sensitivity to perturbations.
To see this, consider a layer with only two neurons, with activations $a_1$ and $a_2$.  Defining $\Delta a = a_1 - a_2$ as the difference in activations, the softmax outputs reduce to sigmoids:
\vspace{-0.2cm}
$$
z_1 = \sigma_1 (t \ba)  = f(- t \Delta a), ~~ z_2 = \sigma_2 (t \ba ) = f( t \Delta a)
$$
where $f(x) = 1/(1+ e^{-x})$ is the standard single-argument sigmoid function.
Since $f(x) \rightarrow 0$ as $x \rightarrow - \infty$, the derivative of the sigmoid, $f'(x) = f(x) f(-x)$, is small for large $|x|$.
Thus, as we increase the separation between the activations (by increasing $|\Delta a|$), the {\it sensitivity} of the softmax output to perturbations decreases. This is in contrast to the ReLU nonlinearity, where perturbations can ride on top of activations in the linear region. 
\vspace{-0.2cm}
\begin{remark}
Even in this simplified setting of two neurons, note that the TEXP inference layer is different from a classical sigmoid nonlinearity.  The TEXP inference output depends on the sigmoid of the {\it difference} in activations, instead of on individual activations as in a classical setting.
\end{remark}
\begin{remark}
As the tilt parameter increases, the sensitivity to perturbations decreases, but this may come at the cost of excessive loss of information due to suppression of weaker activations.
\end{remark}
\begin{remark}
The suppression of small activations by larger ones via softmax leads to a sparse code with a small fraction of strong activations. It is beneficial to further threshold the softmax layer output to zero out small entries, further increasing sparsity and reducing the effective number of dimensions available for perturbations to propagate up.
\end{remark}

\vspace{-0.1cm}
\subsection{Geometric insight into TEXP learning.}\vspace{-0.2cm}
\label{sec:geom_insights}
While our evaluations in Sec.~\ref{sec:experiments} focus on TEXP as a supplement to empirical risk minimization for supervised learning, to obtain geometric insight, we consider unsupervised TEXP training on a single-layer network for two simplified data models, each corresponding to a 2-dimensional {\em signal subspace} embedded in an ambient dimension $d \gg 2$. We apply TEXP training on $M$ randomly initialized neurons $\{\bw_i\}_{i \in [M]}$ (where $M \gg 2$), and ask if we learn neurons aligned with the {\em important} directions in the input distribution. Since the neurons are randomly initialized, we can assume, without loss of generality, that
the first 2 elements of the standard basis, $\mathbf{e}_1= [1,0,\ldots, 0]$ and $\mathbf{e}_2= [0,1,0,\ldots, 0]$, span the 2-dimensional signal subspace.

\begin{figure}[t]
  \centering \vspace{-0.3cm}
  \subfigure[TEXP]{\includegraphics[width=0.49\columnwidth]{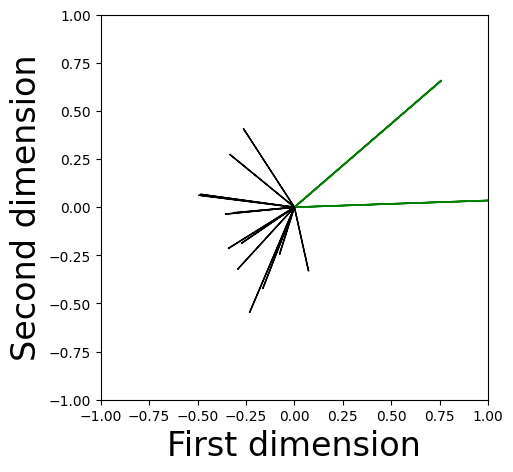}\label{fig:model1_hebb}}
  \hfill
  \subfigure[Balanced TEXP]{\includegraphics[width=0.49\columnwidth]{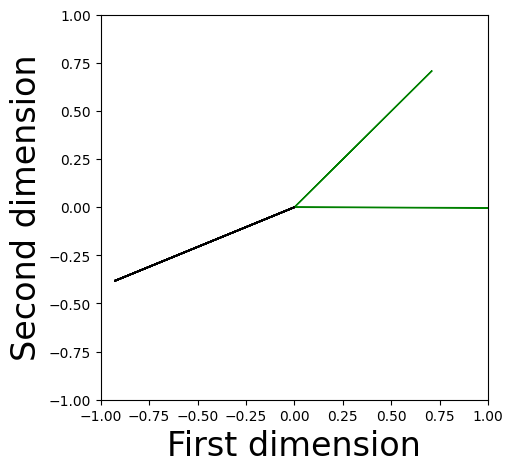}\label{fig:model1_antihebb}}
  \vspace{-0.3cm}
  \caption{Projection of learnt neurons in signal subspace for model 1}
  \label{fig:model_1_neurons}
  \vspace{-0.3cm}
\end{figure}

\paragraph{\bf Model 1:} The input is drawn from a two-component Gaussian mixture, corresponding to one of two equiprobable signals, $\bs_1$ and $\bs_2$, corrupted by white Gaussian noise of variance $\sigma^2$ per dimension.  The input distribution is therefore
given by\vspace{-0.2cm}
\EQ \label{toy_model1}
p( \bx ) = \frac{1}{2} {\cal{N}} ( \bx | \bs_1 , \sigma^2 \bI) + \frac{1}{2} {\cal{N}} ( \bx | \bs_2 , \sigma^2 \bI) \vspace{-0.1cm}
\EN
where $\bx \in {\cal{R}}^d$. This is exactly aligned with our communication-theoretic formulation with two hypotheses. We hope that we learn one or more neurons that align with each of the signal directions $\bs_1$ and $\bs_2$, and that outputs corresponding to the remaining non-useful neurons are small enough that they can be suppressed via TEXP inference. Additionally, when we employ the balanced TEXP objective, the inactive neurons are expected to be rotated away from the signal directions.  We consider an example with signals $\bs_1 = [1,0,\ldots, 0]$, $\bs_2 = [1/\sqrt{2}, 1/\sqrt{2}, 0,\ldots, 0]$, %
$d=10$, $M=20$, and we plot in Fig.~\ref{fig:model1_hebb} the projections, of the learnt neurons, onto the signal space. As expected, TEXP leads to some {\em useful} neurons (shown in green) that align with the signal directions, while the inner products between the remaining {\em non-useful} or  {\em spurious} neurons (shown in black) and the signals are either negative or are small positive numbers (and hence would be significantly attenuated by the softmax in TEXP inference relative to the activations of the useful neurons). The projections of the non-useful neurons into the signal space are of various lengths; that is, they may have substantial energy orthogonal to the signal space. With the {\em balanced} TEXP objective, on the other hand, the non-useful neurons (in black, Fig.~\ref{fig:model1_antihebb}) are rotated away from the signal directions, so that their inner products with signals are negative. Further, the energy of the non-useful neurons is also concentrated in the signal space.

\begin{figure}[t]
  \centering 
  \subfigure[]{\includegraphics[width=0.49\columnwidth]{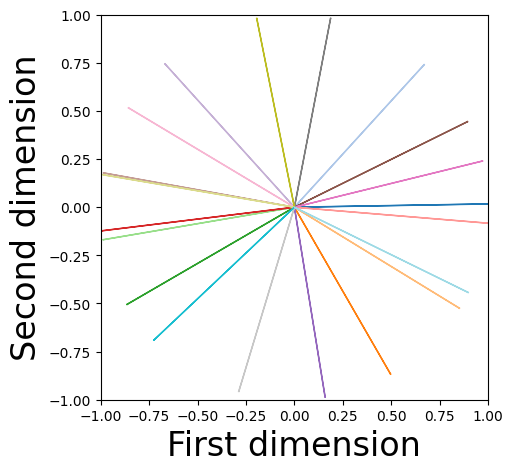}\label{fig:model2_hebb_neurons}}
  \hfill
  \subfigure[]{\includegraphics[width=0.49\columnwidth]{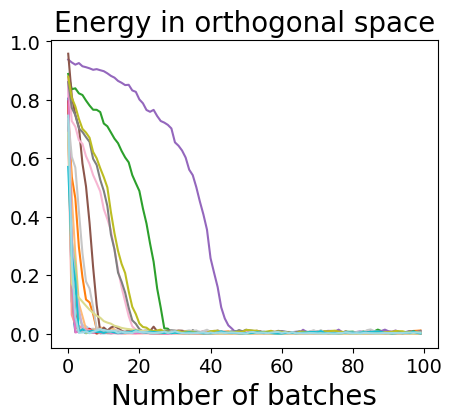}\label{fig:model2_orth_en}}
  \vspace{-0.3cm}
  \caption{Convergence of neurons for model 2 and their respective energies in the orthogonal space.}
  \label{fig:model_2}
  \vspace{-0.4cm}
\end{figure}

\paragraph{\bf Model 2:} The input is a zero mean Gaussian random vector with density $p( \bx ) = {\cal{N}} ( \bx |  \bzero, \bC)$, with eigenvectors taken to be aligned with
the standard basis without loss of generality, and the first two dominant eigen-directions defining the signal subspace:\vspace{-0.2cm}
\EQ \label{covariance_toy_model2} 
\bC =  {\rm diag} \left( A_1^2 + \sigma^2, A_2^2 + \sigma^2, \sigma^2,...,\sigma^2 \right)\vspace{-0.1cm}
\EN
where $A_1^2, A_2^2$ are {\em signal powers} in the dominant eigen-directions, and $\sigma^2$ is the ambient noise variance per dimension.  We can rewrite the model as a random Gaussian signal in a two-dimensional subspace corrupted by white Gaussian noise of variance $\sigma^2$ per dimension:
$\bx = \bs + \bN$, where $\bs = A_1 Z_1 \be_1 + A_2 Z_2 \be_2$ with $Z_1, Z_2$ i.i.d. $\mathcal{N}(0,1)$, and $\bN \sim {\cal{N}}(\bzero, \sigma^2 \bI)$ independent of $Z_1, Z_2$.
Since $(Z_1,Z_2)$ takes a continuum of values in two dimensions, we would need a continuum of signal hypotheses to fit this into our communication-theoretic formulation.  
Thus, given $M \gg 2$ neurons, we expect TEXP training to utilize all available neurons to obtain a sparse code for the random signal based on appropriately quantizing the
continuum of directions that $\bs$ can take in the two-dimensional space. {Fig.~\ref{fig:model2_hebb_neurons}, where we plot the projections of all $M=20$ neurons (in distinct colors) onto the signal space, shows that the energy of the learnt neurons is indeed concentrated in the signal subspace, with more neurons aligned more closely along the first dimension (since $A_1 > A_2$; we set $A_1=3$, $A_2=2$).  Fig.~\ref{fig:model2_orth_en} plots the orthogonal components of the energies of each of these neurons (in respective colors) on the Y-axis, and shows how the energy orthogonal to
the signal subspace dies down quickly during training.} %

The contrasting behavior with these simplified data models indicates how TEXP learning adapts to the richness of the signal subspace to create a sparse code.  
More insights on the sparsity induced by TEXP are deferred to Appendix~\ref{app:toy_models}.

\section{TEXP as a neural network layer}
\label{sec:texp_dnn}\vspace{-0.2cm}
We now translate these ideas to training a convolutional layer in a CNN (the description specializes in a straightforward manner to a fully connected layer).

\paragraph{TEXP inference layer.} We replace a conventional ReLU and batchnorm in the first layer of a neural network by a {\em tilted softmax} and {\em thresholding layer}, and supplement the end-to-end training objective with the TEXP objective for learning matched filters. Our exposition focuses on replacing the first layer by a TEXP layer, but in principle this could be applied to multiple layers.

{\it Convolution with implicit normalization:}  %
For a CNN layer with input $\bx$, we denote the parameters by $\mathcal{W} = \{\bw_i\}_{i \in [M]}$ (where $M$ denotes the number of output channels or the number of filters), with each filter $\bw_i$ being a $k \times k$ kernel with $c_{\text{in}}$ input channels. Thus the dimension of $\bw_i$ is $D = k \times k \times c_{\text{in}}$. Assuming $k$ is odd, consider $\bx(l)$ to be a  $D$-dimensional patch of the input, centered around the spatial location $l$, where $l \in [L]$ and $L$ is the total number of such patches which are convolved with the filters, which depends on the dimensions of the input and the striding and padding. For example, for CIFAR-10 images fed to a VGG-16 model, the first convolution block consists of $M =64$ filters, each a $3\times3$ kernel with stride and padding of $1$. Thus, we have $L = 32\times32=1024$ spatial locations and corresponding input patches. Similarly to~\cite{cekic}, we implicitly normalize the convolution filter weights to unit $\ell_2$ norm, leading to the following convolution output produced at spatial location $l$ due to the $i$-th filter, computed as a tensor inner product as follows: %
\begin{equation} \vspace{-0.3cm}
y_i (l) =  \frac{ (\bx (l))^T \bw_i}{\|\bw_i\|_2}.
\label{eqn:norm_ip}\vspace{0cm}
\end{equation}

{\it Tilted softmax:} Post the convolution, we pass the convolution outputs at each location $l$ through a Tilted Softmax (TS) to obtain {\em posterior probabilities}
\vspace{-0.1cm}
$$
p_i(l) = \frac{\exp({t_{\text{inf}}  y_i(l)})}{\sum_{j = 1}^M \exp({t_{\text{inf}}  y_j(l)})} 
= \sigma_i(t_\text{inf} \by(l)),
$$
where $\by(l) = \{y_i(l), i=1,2,\ldots,M\}$, $t_{\text{inf}}$ is the tilt parameter.  This enforces competition across filters at each location $l$.

{\it Thresholding:} A filter-specific data-adaptive thresholding is performed to obtain TEXP layer outputs:  
\vspace{-0.3cm}
\begin{equation}
o_i(l) =
\left\{
	\begin{array}{ll}
		p_i(l)  & \mbox{if } p_i(l) \geq \tau_i \\
		0 & \text{otherwise}
	\end{array}
\right. 
\end{equation}
This idea of adaptive thresholding is borrowed from ~\cite{cekic}, where $\tau_i$ was set such that a certain fraction (e.g., 10 or 20\%) of the outputs are activated for each filter for any given input. However, we avoid the sorting required for the latter (thus significantly reducing complexity) by setting $\tau_i = m_i + c \sigma_i$, where
$m_i = \frac{1}{L} \sum_{l \in [L]} o_i(l)$ is the mean of the softmax outputs for filter $i$, $\sigma_i$ is corresponding the standard deviation, and $c$ is a hyperparameter (we set $c=0.5$ in our experiments).

\begin{figure}[t]
    \centering \vspace{-0.3cm}
    \includegraphics[width=0.5\columnwidth]{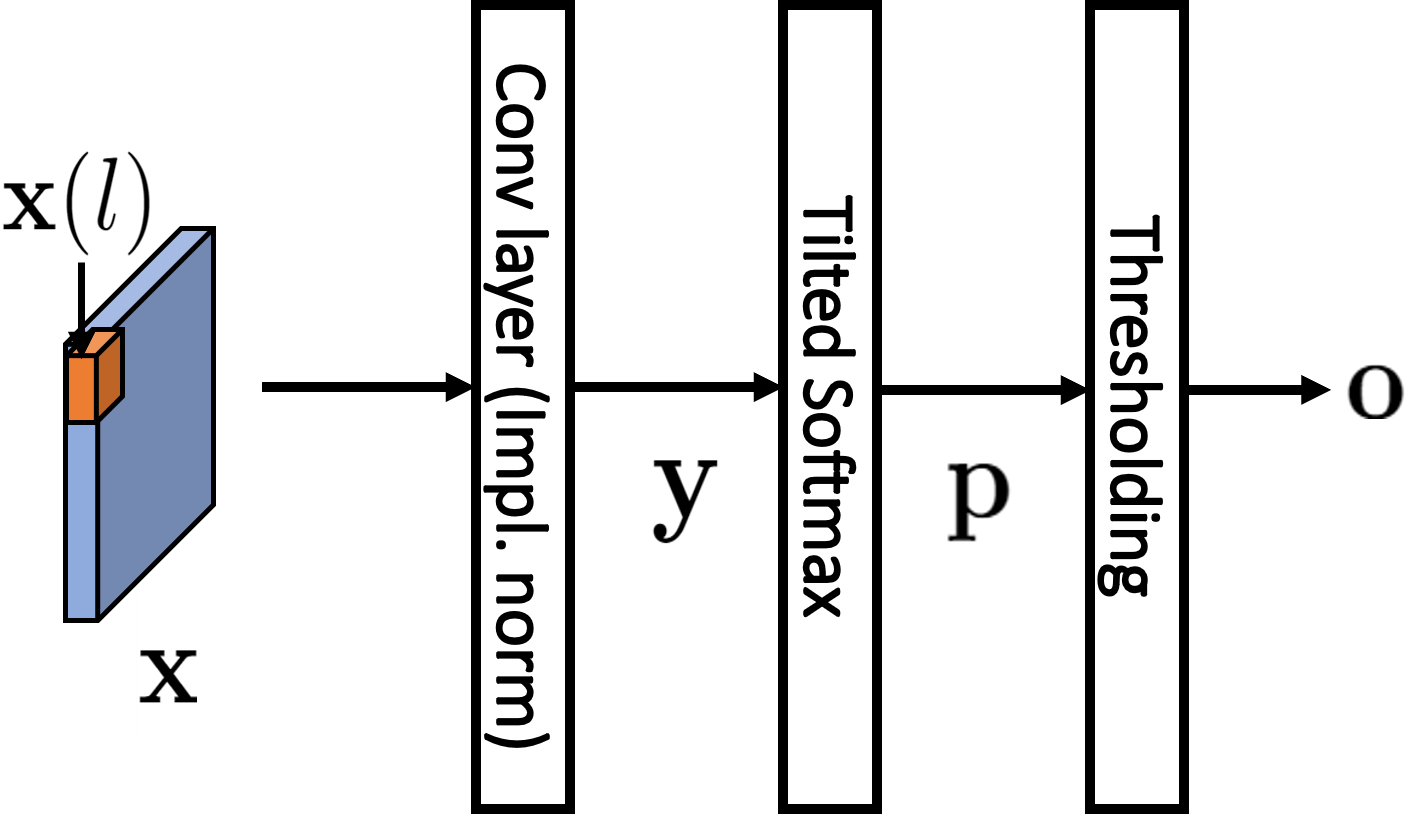}
    \caption{The illustration of a TEXP layer.}
    \label{fig:texp_layer}\vspace{-0.4cm}
\end{figure}

\paragraph{TEXP objective.}
The TEXP objective for a given layer contributed by each training data point is: 
\vspace{-0.2cm}
\begin{equation}
    \mathcal{L}_{\text{TEXP}} = \frac{1}{L} \sum_{l = 1}^{L} \frac{1}{t} \log\left(\frac{1}{M} \sum_{i=1}^{M} \exp(t y_i(l)) \right)
    \label{eqn:texp_obj}\vspace{-0.1cm}
\end{equation}
where $t$ is the training tilt parameter, chosen to be at least as large as the inference tilt $t_{\text{inf}}$ when learning with clean data. The overall cost function takes the form 
\vspace{-0.2cm} $$ \mathcal{L} = \mathcal{L}_{\text{CE}} - \alpha \mathcal{L}_{\text{TEXP}},$$  %
where $\mathcal{L}_{\text{CE}}$ is the standard cross-entropy loss and $\alpha > 0$ is a hyperparameter that determines the relative importance of the TEXP objective compared to the discriminative cross-entropy loss.

\vspace{-.1in}
\section{Experimental setup} \label{sec:experiments}
\vspace{-0.2cm}

We first present details on the baseline and TEXP models employed in the experimental evaluation, the training mechanism and the test metrics. We then report on experiments demonstrating enhancements of robustness for supervised learning of CNNs, replacing the first DNN layer of a baseline architecture with a TEXP layer. Our primary focus is on the CIFAR-10 standard and corruption datasets with the VGG-16 model as the baseline architecture, where we expect significant gains in robustness to noise and other common corruptions in the TEXP models.
The code for all our experiments is available at~\url{https://github.com/bhagyapuranik/texp_for_robustness}.%

\paragraph{Baselines.} We obtain baseline VGG-16 models (with implicit weight normalization), with standard training (which is not expected to be robust), with OOD data augmentation techniques such as AugMix, RandAugment, AutoAugment, and with PGD-based adversarial training~\citep{madry2017towards} with $\ell_{\infty}$ perturbations of budget $\epsilon=2/255$ (which is expected to be robust against a number of other corruptions as well~\citep{yi2021improved}). The HaH model~\citep{cekic} is also used as a baseline for OOD robustness. Like our approach, it supplements training with layer-wise costs. The HaH model modifies $6$ layers, but as we shall see, the TEXP approach with a single layer outperforms it.

\paragraph{Our models.} We modify the first layer of the VGG-16 to a TEXP layer and refer to the model as TEXP-VGG-16. We simplify hyperparameter search for tilt parameters via the following scaling arguments. In view of the implicit normalization of the filters, activations scale with the $\ell_2$ norm of the input $\|\bx\|_2$ to the filter, so that the tilt parameter should be chosen to compensate for this scaling. Making the simplifying assumption that input components are uncorrelated, the energies of the input components add up, and we may assume that $\|\bx\|_2$ scales as $\sqrt{D}$, where $D$ is the dimension of the filter tensor (e.g., for the input of VGG-16, $D=3\times3\times3 = 27$).  Instead of an extensive hyperparameter search, we set $t_{\text{inf}} = 1/\sqrt{D}=0.192$, which results in relatively ``soft'' decisions at the softmax output.  For the clean training data, we set $t = 10/\sqrt{D} = 1.92$, so TEXP training makes ``harder'' decisions in favor of winners when learning signal templates.  The relative weight of the TEXP objective (\ref{eqn:texp_obj}) in comparison to the cross-entropy loss is set to $\alpha=0.001$, chosen so that the magnitude of the weighted TEXP objective is smaller than the cross-entropy loss. We use the same parameters when supplementing TEXP with data augmentation techniques and adversarial training with $\ell_{\infty}$ perturbations of budget $\epsilon=2/255$ (better performance may be obtained by further fine-tuning TEXP parameters for each setting). We provide a detailed study on the sensitivity of the TEXP models to the different TEXP parameters in Appendix~\ref{app:hyp_sense}.
{We also report on an alternate, computationally intensive variant of TEXP, in Appendix~\ref{app:expensive_texp_variant}}.

\paragraph{Training.} For all VGG-16 based models, we employ the ADAM optimizer~\citep{kingma2014adam} with a multi-step learning rate, beginning with $0.001$, and decreasing by a factor of $10$ at epochs $60$ and $80$. We train the models for $100$ epochs.

\begin{table*}[t!]
\centering\vspace{-0.4cm}
\resizebox{\textwidth}{!}{
\begin{tabular}{c@{\hspace{4mm}}c@{\hspace{8mm}}c@{\hspace{4mm}}c@{\hspace{4mm}}c@{\hspace{8mm}}c@{\hspace{4mm}}c@{\hspace{4mm}}c@{\hspace{4mm}}}\\
\toprule
Model  &  Clean & \multicolumn{3}{c}{OOD corruptions} & \multicolumn{3}{c}{Adversarial perturbations} \\
\cmidrule(r{23pt}){3-5} \cmidrule(r){6-8}
 & & Noise & {Min/Avg} & {Min/Avg}& {{Autoattack $\ell_1$}}& {{Autoattack $\ell_2$}} & {{Autoattack $\ell_{\infty}$}}\\ 
& &{$\nu=0.1$} & {corruptions} & {severity level:5} & {{adv, $\epsilon=3$}}& {{adv, $\epsilon=0.25$}} & {{adv, $\epsilon=2/255$}}\\ \toprule

VGG-16 & $92.26 \pm 0.04 $ & $24.80 \pm 1.24$  &$46.86\pm 1.26 / 72.28 \pm 0.26$ & $19.56\pm 0.73 / 54.70\pm 0.40$&$10.14\pm 0.22$ &{$13.34\pm 0.14$ } & {$10.30\pm 0.21$}\\
HaH (\cite{cekic}) & $87.72\pm 0.15$ & $62.76\pm 0.40$  & $59.56\pm 0.42/77.02\pm 0.21$& $49.06\pm 0.88/67.80\pm 0.27$& $29.98\pm0.45$&{$26.30\pm 0.52$} & {$20.04\pm 0.38$}\\
TEXP-VGG-16 & $88.28\pm 0.12$ & $75.14\pm 0.20$  & $73.68\pm 0.22/80.40\pm 0.07$ & $52.38\pm 0.81/72.56\pm 0.14$& $46.48\pm0.95$&{$50.90\pm 0.16$} & {$41.50\pm 0.21$}\\
\midrule
VGG-16 + AugMix & $92.98\pm 0.06$ & $62.92\pm 0.74$  &$65.12\pm 0.35/83.58\pm 0.09$ & $42.12\pm 0.79/74.00\pm 0.16$&$17.88\pm0.26$ &{$18.16\pm 0.15$ } & {$13.60\pm 0.17$}\\
TEXP-VGG-16 + AugMix & $88.84\pm 0.21$ & $78.90\pm 0.04$  & ${77.28\pm 0.20/83.54\pm 0.05}$ & ${62.94\pm 0.53/78.30\pm 0.07}$& $48.68\pm0.52$&{$52.20\pm0.23$} & {$42.52\pm 0.20$}\\
\midrule
VGG-16 + RandAug & $93.32\pm 0.11$ & $43.32\pm 0.72$  &$63.24\pm 0.45/80.68\pm 0.17$ & $39.98\pm 1.01/66.96\pm 0.30$&$19.76\pm0.08$ &{$18.38\pm0.47$ } & {$14.30\pm 0.37$}\\
TEXP-VGG-16 + RandAug & $89.90\pm 0.08$ & $74.26\pm 0.07$  & $75.48\pm 0.09/82.86\pm 0.02$ & $57.52\pm 0.19/75.78\pm 0.07$&$50.16\pm0.24$ &{$50.82\pm 0.24$} & {$40.02\pm0.34$}\\
\midrule
VGG-16 + AutoAug& ${93.50\pm 0.03}$ & $46.54\pm 0.54$  &$59.84\pm 0.52/81.58\pm 0.14$ & $37.08\pm 0.23/70.66\pm 0.18$&$15.66\pm0.19$ &{$13.50\pm 0.23$ } & {$9.78\pm 0.20$}\\
TEXP-VGG-16 + AutoAug & $90.06\pm 0.10$ & $72.66\pm0.46$  & $71.98\pm 0.24/82.58\pm 0.12$ & $54.14\pm 0.89/75.50\pm 0.18$&$47.62\pm0.16$ &{$46.96\pm 0.31$} & {$35.00\pm 0.32$}\\
\midrule
VGG-16 + Adv Tr& $88.04\pm 0.12$ & $78.78\pm 0.45$  & $50.52\pm0.66/79.44\pm 0.12$ & $17.60\pm 0.39/70.64\pm 0.13$ &$51.26\pm0.71$ &{$72.60\pm 0.23$ } & {${72.82\pm 0.23}$}\\
TEXP-VGG-16 + Adv Tr &$86.38\pm 0.07$ & ${81.08\pm 0.28}$  & $67.72\pm 0.73/80.38\pm 0.14$ & $37.08\pm0.85/74.02\pm 0.22$&$64.50\pm0.45$ &{$71.02\pm 0.40$} & {$66.76\pm0.29$}\\
\bottomrule
\end{tabular}}
\caption{Enhanced robustness to corruptions and mild adversarial attacks under VGG-16 based TEXP models on CIFAR-10 clean and corruptions datasets. All numbers reported are average test accuracies $\pm$ standard error.
}
\label{tab:performance}
\end{table*}

\begin{table*}[t!]
\centering
\resizebox{\textwidth}{!}{
\begin{tabular}{c@{\hspace{2mm}}c@{\hspace{2mm}}c@{\hspace{2mm}}c@{\hspace{2mm}}c@{\hspace{2mm}}c@{\hspace{2mm}}c@{\hspace{2mm}}c@{\hspace{2mm}}c@{\hspace{2mm}}c@{\hspace{2mm}}c@{\hspace{2mm}}c@{\hspace{2mm}}c@{\hspace{2mm}}c@{\hspace{2mm}}c@{\hspace{2mm}}c@{\hspace{2mm}}c@{\hspace{2mm}}c@{\hspace{2mm}}c@{\hspace{2mm}}c@{\hspace{2mm}}}

\toprule

Corruptions $\rightarrow$ &  \multicolumn{4}{c}{Noise} & \multicolumn{5}{c}{Weather} & \multicolumn{5}{c}{Blur} & \multicolumn{5}{c}{Digital} \\ 
\cmidrule(r){2-5} \cmidrule(r){6-10} \cmidrule(r){11-15} \cmidrule(r){16-20}
Models $\downarrow$ & Gauss. & Shot & Speck. & Imp. & Snow & Frost & Fog & Brig. & Spat. & Defoc. & Gauss. & Glass&Motion & Zoom & Cont. & Elas. & Pixel. &JPEG&Satur. \\ \midrule

VGG-16 & $24.3$ & $31.8$ & $38.4$ & $19.1$ & $73.3$ & $62.0$ & ${63.8}$ & ${87.9}$ & $67.3$ & $50.8$ & $39.8$ & $47.6$ & $60.0$ & $61.5$ & $19.9$ & $75.6$ & $54.6$ & $77.4$ & $82.4$\\

HaH~\citep{cekic} & $61.7$ & $61.7$ & $59.2$ & $46.3$ & $73.8$ & $72.3$ & $62.8$ & $83.2$ & $76.7$ & $64.3$ & $58.4$ & $53.2$ & $65.1$ & $68.9$ & $76.0$ & $74.0$ & $60.5$ & $79.3$ & $79.6$\\

TEXP-VGG-16 & $75.3$ & $76.5$ & $75.5$& $61.3$& $76.4$&$76.8$& $51.8$& $83.2$& $76.1$& $68.9$ & $63.4$ & $68.6$ & $65.0$ & $74.2$ & $66.0$&  $75.2$& $80.8$& $82.9$& $78.8$\\
\midrule

VGG-16 + AugMix& $60.7$&  $68.1$&  $71.3$&  $44.9$&  $80.2$&  $75.3$&  $76.5$&  $89.7$&  $81.7$&  ${84.8}$&  ${80.8}$&  $59.6$&  ${81.4}$&  ${84.0}$&  $40.0$&  $79.5$&  $69.4$&  $82.0$&  $86.9$\\

TEXP-VGG-16 + AugMix& $78.9$& $79.5$& $79.0$& $67.7$& $78.4$& ${79.0}$& $62.2$& $83.8$& $78.8$& $81.5$& $79.8$& $72.4$& $77.1$& $82.6$& $75.5$& $78.6$& ${83.6}$& $83.7$& $81.6$\\
\midrule

VGG-16 + RandAug& $44.7$&  $53.5$&  $57.5$&  $40.0$&  $78.6$&  $72.8$&  $71.0$&  $90.9$&  ${85.3}$&  $63.6$&  $52.9$&  $61.0$&  $67.8$&  $71.7$&  $48.3$&  ${79.9}$&  $56.9$&  $81.7$&  $88.5$\\

TEXP-VGG-16 + RandAug & $74.1$& $75.1$& $72.7$& $57.1$& ${79.1}$& $78.7$& $60.3$& $88.6$& $81.3$& $73.4$& $68.7$& $70.7$& $70.8$& $77.4$& $83.5$& $78.3$& $79.4$& $84.5$& $85.8$\\
\midrule

VGG-16 + AutoAug& $45.7$&  $53.1$&  $56.7$&  $37.1$&  $77.2$&  $69.8$&  ${81.1}$&  ${91.9}$&  $81.1$&  $79.1$&  $75.2$&  $51.8$&  $75.2$&  $81.1$&  $80.0$&  $76.5$&  $50.4$&  $80.2$&  ${90.2}$\\

TEXP-VGG-16 + AutoAug & $72.3$& $72.5$& $70.8$& $53.1$& $76.9$& $76.1$& $62.9$& $88.3$& $77.5$& $76.1$& $72.9$& $65.6$& $72.4$& $79.8$& ${86.0}$& $76.5$& $77.4$& $84.5$& $86.2$\\
\midrule

VGG-16 + Adv Tr& $79.8$&  $81.1$&  $80.3$&  $62.7$&  $74.3$&  $73.3$&  $33.2$&  $76.8$&  $77.7$&  $71.1$&  $66.8$&  $76.0$&  $69.1$&  $74.9$&  $18.3$&  $78.4$&  $82.6$&  ${84.8}$&  $76.6$\\

TEXP-VGG-16 + Adv Tr &${81.6}$& ${82.3}$& ${81.9}$& ${74.8}$& $71.9$& $75.8$& $39.0$& $76.9$& $78.5$& $75.9$& $72.8$& ${76.8}$& $73.1$& $78.3$& $52.9$& $78.6$& $83.2$& $84.0$& $76.3$\\

\bottomrule

\end{tabular}
}
\caption{Robustness to common corruptions of the highest severity level in the CIFAR-10-C dataset. All numbers reported are test accuracies.} 
\label{table:corruption} \vspace{-0.4cm}
\end{table*}

\paragraph{Evaluation metrics.} We evaluate over $19$ different common corruptions on the CIFAR-10-C~\citep{hendrycks2018benchmarking} dataset. %
We report the test accuracies for minimum (worst-case) and average over all the corruptions, for both the entire dataset comprising of 5 different severity levels, and also on specifically the corruptions of the highest severity.%
We also separately report on the corrupted data formed by the addition of Gaussian noise with standard deviation $\nu=0.1$ (since our TEXP formulation was motivated by hypothesis testing with Gaussian noise, we expect enhanced robustness to Gaussian noise). Finally, we explore whether TEXP (even without adversarial training) enhances resilience against mild adversarial attacks, by evaluating all trained models on different $\ell_p$ attacks such as $\ell_1$ with budget $\epsilon=3$, $\ell_2$ with $\epsilon=0.25$ and $\ell_{\infty}$ with $\epsilon=2/255$ respectively. We use AutoAttack~\citep{croce2020reliable}, suggested by RobustBench~\citep{croce2020robustbench}, which is parameter-free and consists of a suite of different attacks (in particular, we employ the \textit{standard} version composed of APGD-CE, APGD-T, FAB-T and Square attacks). 

TEXP is intended to produce sparse, strong activations; we report on sparsity measures for TEXP layer outputs in Appendix~\ref{app:texp_dnn_details}.

\paragraph{Broader applicability.}
We illustrate the applicability of TEXP to different architectures and larger datasets via preliminary experiments on the CIFAR-100 dataset using Wide-ResNet-28-10~\citep{zagoruyko2016wide} baseline and on ImageNet-1K~\citep{russakovsky2015imagenet} with ResNet50 baseline. For TEXP-WideResNet-28-10, we set the TEXP parameters as $t_{\text{inf}} = 1/\sqrt{D} = 0.192$, $t=4 t_{\text{inf}} = 0.768$ and $\alpha=0.0005$. For TEXP-ResNet-50, we set the TEXP parameters as $t_{\text{inf}} = 8/\sqrt{D} = 0.656$, $t=10 t_{\text{inf}} = 0.656$ and $\alpha=0.01$. For both these ResNet family backbones, we employ SGD optimizer with momentum ($0.9$), initial learning rate of $0.1$, and weight decay. The ResNet-50 models are trained for 90 epochs, with a $10\times$ learning rate reduction every $30$ epochs. The WideResNet models are trained for 200 epochs, with $5\times$ learning rate reduction every $60$ epochs. 

\vspace{-0.3cm}
\section{Experimental results}
\vspace{-0.3cm}
\paragraph{Improvement in robustness against OOD corruptions.} 
In Table~\ref{tab:performance}, we report the enhanced {\em general-purpose} robustness of the TEXP approach on the CIFAR-10 dataset. The reported numbers are test accuracies averaged across five runs, along with the standard error. Please refer to Appendix~\ref{app:stat_sig} for details on the computation of the standard errors. Focusing on the robustness to common corruptions, we can observe that TEXP provides significant gains in robustness to noise and other out-of-distribution (OOD) corruptions (both at the highest severity level and all levels) in comparison to standard VGG and HaH baselines. 
We benefit substantially against the worst corruption (i.e., the one with minimum accuracy among all corruptions). %
TEXP combined with data augmentations and adversarial training provides even more powerful enhancements in OOD robustness, outperforming the backbones, both in terms of average over all kinds of corruptions and the minimum or worst-case among the different corruptions. We observe that TEXP augmented with AugMix provides the best OOD robustness overall.
Table~\ref{table:corruption} reports the robustness of the models to each of the $19$ common corruptions (in the CIFAR-10-C dataset) separately for the highest severity level of $5$, and shows that TEXP models are superior in obtaining robustness against most types of corruptions, and in particular, against various noise corruptions. \edits{The robust accuracies here are reported for a single run.} While vanilla adversarial training helps in robustness to noise, it deteriorates performance against contrast~\citep{yin2019fourier, kireev2022effectiveness, machiraju2022empirical}. This issue is alleviated by TEXP.

\begin{table*}[t!]
\centering\vspace{-0.4cm}
\resizebox{0.7\textwidth}{!}{
\begin{tabular}{c@{\hspace{4mm}}c@{\hspace{4mm}}c@{\hspace{4mm}}c@{\hspace{4mm}}c@{\hspace{4mm}}c@{\hspace{4mm}}c@{\hspace{4mm}}c@{\hspace{4mm}}c@{\hspace{4mm}}}\\
\toprule
Model  &  \multicolumn{2}{c}{Clean} & \multicolumn{2}{c}{Noise} & \multicolumn{2}{c}{Min/Avg} & \multicolumn{2}{c}{Min/Avg} \\
 &  \multicolumn{2}{c}{} & \multicolumn{2}{c}{$\nu=0.1$} & \multicolumn{2}{c}{corruptions} & \multicolumn{2}{c}{severity level:5} \\
\cmidrule(r){2-3} \cmidrule(r){4-5} \cmidrule(r){6-7} \cmidrule(r){8-9}
& Top-1 & Top-5 & Top-1 & Top-5 & Top-1 & Top-5 & Top-1 & Top-5 \\ \toprule
WRN-28-10 (CIFAR-100) &$81.2$&$95.3$&$9.6$&$25.6$&$17.8/51.3$&$35.6/72.9$&$2.9/34.8$&$10.9/57.6$ \\
TEXP-WRN-28-10  &$78.4$&$93.6$&$31.4$&$57.3$&$26.0/56.7$&$52.8/78.7$&$12.4/40.9$&$30.1/65.0$ \\ \midrule
ResNet50 (ImageNet)&$75.5$&$92.6$&$55.6$&$79.1$&$24.3/38.3$&$41.0/58.8$&$3.3/17.9$&$8.9/33.5$\\
TEXP-ResNet-50  &$72.0$&$90.6$&$62.6$&$84.3$&$26.6/41.8$&$45.8/62.5$&$3.3/21.0$& $9.4/37.5$ \\
\bottomrule
\end{tabular}}
\caption{Test accuracy for WRN-28-10 based TEXP model on CIFAR-100 clean and corruptions datasets and ResNet-50 based TEXP model on ImageNet-1K clean and corruptions datasets. 
}
\label{table:perf-cif100} \vspace{-0.4cm}
\end{table*}

\paragraph{Robustness against mild adversarial perturbations.}
\edits{Table~\ref{tab:performance} also shows that TEXP-based models enhance robustness to mild adversarial perturbations even without adversarial training. When TEXP is combined with adversarial training, its adversarial robustness generalizes significantly better than the baseline: for example, adversarial training with $\ell_{\infty}$-bounded attacks yields improved performance against (a very different kind of) $\ell_1$-bounded attack. %

\paragraph{Robustness under different architectures and datasets.}
The performance of the TEXP models is contrasted against baselines for the CIFAR-100 (with WideResNet-28-10) and ImageNet-1k (with ResNet-50) datasets in Table~\ref{table:perf-cif100}. We chose the TEXP parameters for these models through a mild search, settling on a combination that sacrifices clean accuracy by $3-4\%$, as we had for CIFAR-10. The results demonstrate improved OOD robustness despite minimal effort in hyperparameter tuning. We expect that more fine-grained adjustments of tilts combined with replacing multiple deeper layers with TEXP will further enhance performance, but these preliminary results do illustrate the potential gains from applying TEXP to different architectures and larger datasets.

\paragraph{Ablation study and connections to activation pruning.}
To understand the contributions of each of the components of our approach, we  train and evaluate models by cumulatively adding the thresholding, tilted softmax and TEXP objective to baseline VGG. Detailed results are presented in Appendix~\ref{app:abl_study} which show the cumulative benefits through improved robustness to both corruptions and mild adversarial attacks.  An interesting connection, among works following thresholding strategies, is the work on stochastic activation pruning (SAP)~\citep{dhillon2018stochastic}, proposed as an adversarial defense and subsequently broken in~\cite{dhillon2020erratum}. SAP also results in sparse activations, retaining activations at each layer with probabilities proportional to their magnitude, while pruning the rest.  Our deterministic thresholding scheme, used by itself, serves as a proxy for the strategy in SAP. However, the ablations in Appendix~\ref{app:abl_study} show that the following key aspects of our approach not present in SAP are critical in enhancing robustness: (a) the TEXP objective aligns filters to match incoming patterns to promote strong activations, so that activation pruning results in less information loss, (b) TEXP inference employs a softmax nonlinearity which enhances resilience, as discussed in Section \ref{sec:comm_theory_motivation}.  
}

\paragraph{Computation cost.} The average time taken to execute $1$ epoch of training of a VGG-16 baseline is $5.93$ sec, in comparison to $8.23$ sec for TEXP-VGG-16, on an NVIDIA A100 GPU, while inference over the entire CIFAR-10 dataset takes an average of $1.13$ sec for baseline and $1.12$ sec for TEXP. The average training (inference) time over an NVIDIA GeForce 1080 Ti is $18.9$ sec ($2.79$ sec) for baseline VGG and $27.5$ sec ($2.86$ sec) for TEXP. {We acknowledge that our TEXP code is not optimized for compute speed}, and we believe that the training time may be improved. On the other hand, we observe that introducing TEXP layer results in little inference-time overhead to the network.
In contrast to the adaptive thresholding of~\cite{cekic}, we do not require sorting of activations to set the thresholds, significantly improving computation efficiency. Please refer to Appendix~\ref{app:compute} for additional details. %

\vspace{-0.3cm}
\section{Conclusion}
\vspace{-0.3cm}
Our communication-theoretic approach enhances robustness via neuronal competition in representing layer inputs during both training and inference: for a Gaussian model of data noise, the TEXP learning objective is derived as maximum likelihood estimation of signal templates, and TEXP inference as posterior probability computation. Geometric insight via unsupervised learning on simplified models illustrates how TEXP adapts to the richness of the set of possible inputs, while supplementing supervised learning in CNNs with TEXP training and inference at the input layer is shown to yield gains in OOD robustness for image datasets.  Extensive experiments on a VGG model for CIFAR-10 demonstrate that, in addition to robustness gains with TEXP alone, our approach also combines well with data augmentation strategies.  Preliminary results with a TEXP input layer for ResNet architectures for CIFAR-100 and ImageNet also demonstrate gains in robustness, indicating the promise of this approach for a variety of datasets and architectures. We hope that our results stimulate further work in this area, including interesting questions regarding hyperparameter optimization and training approaches for multiple TEXP layers, and addressing robustness against strong adversarial attacks in addition to OOD robustness. { Furthermore, the recipes of TEXP appear to have similarities with elements of transformer architecture, such as matching queries with keys, and softmax computations. We plan to investigate these connections.}

\vspace{-0.2cm}
\section{Broader impact and limitations}
\vspace{-0.3cm}
\label{app:broad_impact_limit}
The existing techniques for improving robustness are mostly through the application of data augmentations, changing the optimization loss function, or both. Our approach is complementary, focusing on matching the signal to the input layer of the network, which aligns with the broader goal of making deep networks more interpretable as well. Our experimental results confirm that our method indeed complements  standard data augmentation techniques, thereby expanding its applicability to various tasks. A limitation of our work in the current form is that we are yet to develop concrete design guidelines for setting the tilt parameters, which is useful for optimizing the performance of our approach for different architectures. We also note that adding an additional loss function for training results in an increase of the computational cost of training. On the other hand, the computational cost of inference is not impacted substantially, TEXP inference via softmax and thresholding is not significantly more complex than standard ReLU and batch norm. In the future, we will focus on maximizing the effectiveness of our approach across different datasets and models, to ascertain the generalizability. Nonetheless, our findings underscore the value of the tilted exponential layer in enhancing robustness to OOD corruptions, which is important in many practical machine learning tasks where test samples in the real-world are often different from the curated training data.

\vspace{-0.2cm}
\section*{Acknowledgements} \vspace{-0.2cm}
We thank the anonymous reviewers for their valuable suggestions. The work of BP and UM was supported in part by the National Science Foundation under Grants CIF-1909320 and CIF-2224263.

\bibliography{references, tilt_ahmad}

\clearpage
\section*{Checklist}

 \begin{enumerate}

 \item For all models and algorithms presented, check if you include:
 \begin{enumerate}
   \item A clear description of the mathematical setting, assumptions, algorithm, and/or model. \\ {\bf Yes}
   \item An analysis of the properties and complexity (time, space, sample size) of any algorithm. \\
   {\bf Yes}, the training and inference times of our model are compared with baselines.
   \item (Optional) Anonymized source code, with specification of all dependencies, including external libraries. \\
   {\bf Yes}, the source code is available in the repository referenced in Section~\ref{sec:experiments}
 \end{enumerate}

 \item For any theoretical claim, check if you include:
 \begin{enumerate}
   \item Statements of the full set of assumptions of all theoretical results. \\
   {\bf Not Applicable}
   \item Complete proofs of all theoretical results. \\
   {\bf Not Applicable}
   \item Clear explanations of any assumptions. \\
   {\bf Not Applicable}
 \end{enumerate}

 \item For all figures and tables that present empirical results, check if you include:
 \begin{enumerate}
   \item The code, data, and instructions needed to reproduce the main experimental results (either in the supplemental material or as a URL). \\
   {\bf Yes}, the code is provided, along with instructions to reproduce main results.
   \item All the training details (e.g., data splits, hyperparameters, how they were chosen). \\
   {\bf Yes}, please refer to the paragraph on ``Our models'' in Section~\ref{sec:experiments}.
    \item A clear definition of the specific measure or statistics and error bars (e.g., with respect to the random seed after running experiments multiple times). \\
    {\bf Yes}, please refer to the paragraph on ``Evaluation metrics'' in Section~\ref{sec:experiments} for the metrics used in establishing robustness, and Section~\ref{app:stat_sig} for statistical significance of the empirical results. 
    \item A description of the computing infrastructure used. (e.g., type of GPUs, internal cluster, or cloud provider). \\
    {\bf Yes}, provided in supplementary materials, Section~\ref{app:compute}. 
 \end{enumerate}

 \item If you are using existing assets (e.g., code, data, models) or curating/releasing new assets, check if you include:
 \begin{enumerate}
   \item Citations of the creator If your work uses existing assets. \\
   {\bf Yes}, standard open-source image datasets used are cited.
   \item The license information of the assets, if applicable. \\
   {\bf Not Applicable}
   \item New assets either in the supplemental material or as a URL, if applicable. \\
   {\bf Not Applicable}
   \item Information about consent from data providers/curators. \\
   {\bf Not Applicable}
   \item Discussion of sensible content if applicable, e.g., personally identifiable information or offensive content. \\
   {\bf Not Applicable.}
 \end{enumerate}

 \item If you used crowdsourcing or conducted research with human subjects, check if you include:
 \begin{enumerate}
   \item The full text of instructions given to participants and screenshots. \\
   {\bf Not Applicable}
   \item Descriptions of potential participant risks, with links to Institutional Review Board (IRB) approvals if applicable. \\
   {\bf Not Applicable}
   \item The estimated hourly wage paid to participants and the total amount spent on participant compensation. \\
   {\bf Not Applicable}
 \end{enumerate}

 \end{enumerate}

\onecolumn
\appendix

\section*{\LARGE Appendix}
\definecolor{ruddy}{rgb}{0.0, 0.0, 0.0}

\addtocontents{toc}{\protect\setcounter{tocdepth}{2}}
\vspace{0.5cm}
\noindent{\bfseries{\Large Contents}\hfill Page No.\vspace \bigskipamount \par }
\setcounter {tocdepth}{2}
\contentsline {section}{\numberline {A}Geometric insights into TEXP: details on the simplified data models}{14}{appendix.A}%
\contentsline {section}{\numberline {B}Sensitivity to TEXP parameters}{15}{appendix.B}%
\contentsline {section}{\numberline {C}TEXP variant with expanded neural competition }{16}{appendix.C}%
\contentsline {section}{\numberline {D}Sparsity of activations after first DNN layer}{17}{appendix.D}%
\contentsline {section}{\numberline {E}Statistical significance of the results}{18}{appendix.E}%
\contentsline {section}{\numberline {F}Ablation study}{18}{appendix.F}%
\contentsline {section}{\numberline {G}Computation time}{18}{appendix.G}%

\definecolor{ruddy}{rgb}{1.0, 0.0, 0.16}

\clearpage

\section{Geometric insights into TEXP: details on the simplified data models}
\label{app:toy_models}
Recall the data model $1$ in Sec.~\ref{sec:comm_theory_motivation}, where the input was drawn from
\begin{equation*}
  p( \bx ) = \frac{1}{2} {\cal{N}} ( \bx | \bs_1 , \sigma^2 \bI) + \frac{1}{2} {\cal{N}} ( \bx | \bs_2 , \sigma^2 \bI)   
\end{equation*}
with signals $\bs_1 = [1,0,\ldots, 0]$, $\bs_2 = [1/\sqrt{2}, 1/\sqrt{2}, 0,\ldots, 0]$, $d=10$ and $M=20$. Under the TEXP objective, depending upon the initializations of the neurons, one or more neurons could align with each of the two signal directions, leading to several ``useful" neurons (although not apparent in Fig.~\ref{fig:model1_hebb}, there are multiple useful neurons in the directions of $\bs_1$ and $\bs_2$). The activations produced by useful neurons, which are large, survive through the tilted softmax, while the rest are expected to be attenuated. 

\begin{figure}[h]
  \centering 
  \subfigure[Histogram of activations at the output of linear layer]{\includegraphics[width=0.35\columnwidth]{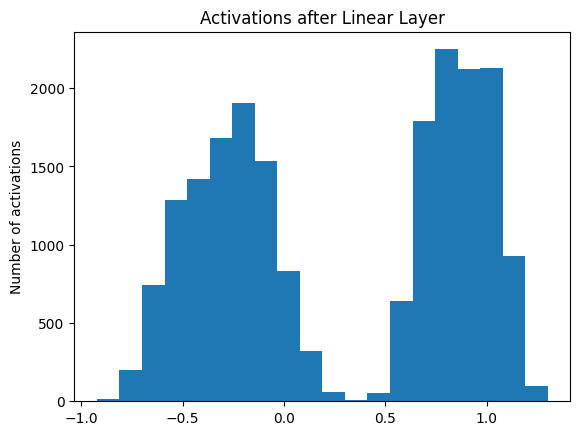}\label{fig:x}}
  \hspace{0.5cm}
  \subfigure[Histogram of activations after passing through a tilted softmax with $t_{\text{inf}} = 1,3$.]{\includegraphics[width=0.35\columnwidth]{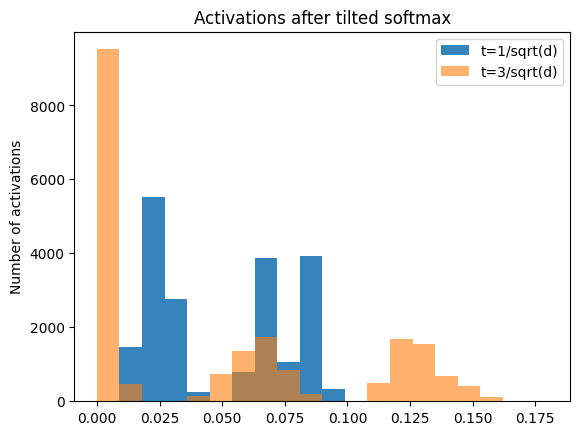}\label{fig:xx}}
  \caption{Simplified data model 1, under TEXP objective}
  \label{fig:mod1_act_hist}
\end{figure}

Focusing on the TEXP objective, we show in Fig.~\ref{fig:mod1_act_hist} the histograms of the activations (i) at the output of the linear layer (ii) after passing through tilted softmax for two differnet values of $t_{\text{inf}} = 1,3$, while the training tilt parameter is $t=10$. We can observe how a stronger tilted softmax can polarize the activations more strongly. In addition, since there could be multiple useful neurons aligned with a signal direction, we could further prune similar neurons, to achieve a more sparse output.

\begin{figure}[h]
  \centering 
  \subfigure[Histogram of activations at the output of linear layer]{\includegraphics[width=0.35\columnwidth]{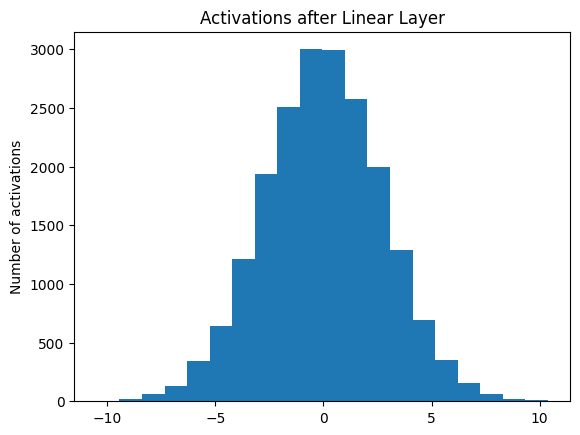}\label{fig:y}}
  \hspace{0.5cm}
  \subfigure[Histogram of activations after passing through a tilted softmax with $t_{\text{inf}} = 1,3$.]{\includegraphics[width=0.35\columnwidth]{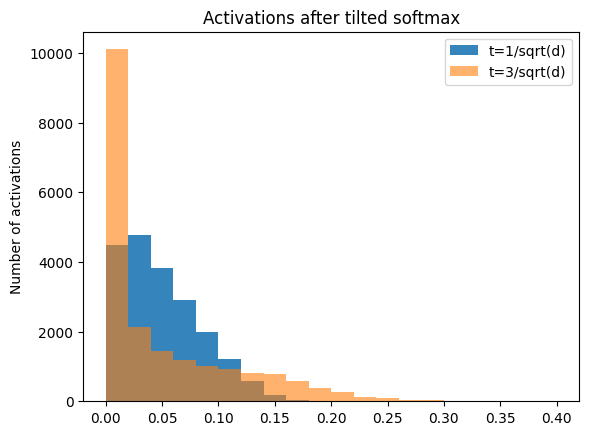}\label{fig:yy}}
  \caption{Simplified data model 2, under TEXP objective}
  \label{fig:mod2_act_hist}
\end{figure}

Similarly, for data model 2, the histograms of the activations at the output of the linear layer, and post tilted softmax are shown in Fig.~\ref{fig:mod2_act_hist}. Thus we can observe that TEXP approach encourages sparse activations.

\section{Sensitivity to TEXP parameters}
\label{app:hyp_sense}
\begin{figure}[b!]
  \centering 
  \subfigure[Average corruption accuracy over all severities]{\includegraphics[width=0.39\columnwidth]{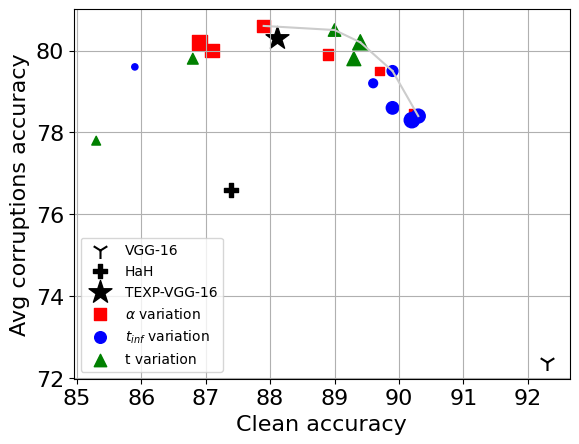}\label{fig:z}}
  \hspace{0.5cm}
  \subfigure[Average (severity level 5) corruption accuracy]{\includegraphics[width=0.4\columnwidth]{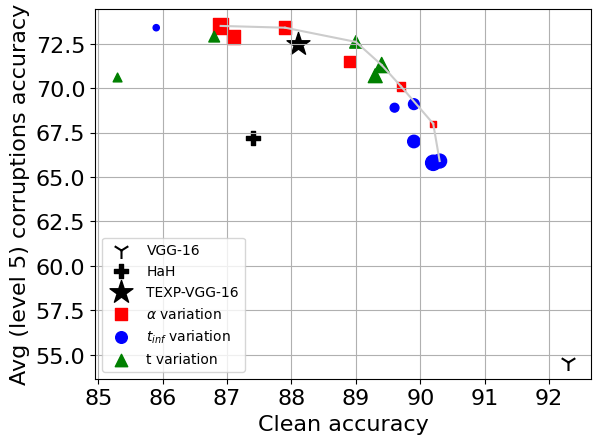}\label{fig:zz}}
  \caption{Robustness-accuracy trade-offs obtained by varying the TEXP parameters. Increasing marker sizes represent increasing values of the corresponding TEXP parameter. The red markers with increasing sizes encode $\alpha$ variation for $\alpha \in$ $[0.00001, 0.0001, 0.0005, 0.002, 0.005, 0.01]$. The blue markers with increasing sizes encode $t_{\text{inf}}$ variation, for $t_{\text{inf}}\in $ [$0.5/\sqrt{D}$, $2/\sqrt{D}$, $3/\sqrt{D}$, $4/\sqrt{D}$, $8/\sqrt{D}$,  $16/\sqrt{D}$]. The green markers with increasing sizes encode $t = \beta t_{\text{inf}}$ variation for $\beta \in $ $[1, 5, 15, 25, 50]$.}
  \label{fig:rob_acc}
\end{figure}
\begin{figure}[b!]
  \centering 
  \subfigure[Minimum corruption accuracy (all severities)]{\includegraphics[width=0.38\columnwidth]{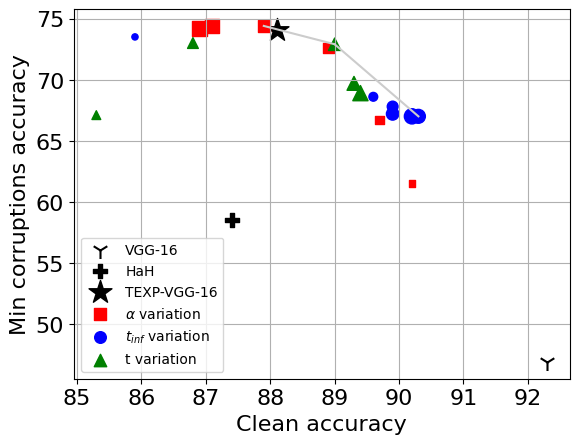}\label{fig:z_min}}
  \hspace{0.5cm}
  \subfigure[Minimum corruption accuracy for highest severity level]{\includegraphics[width=0.38\columnwidth]{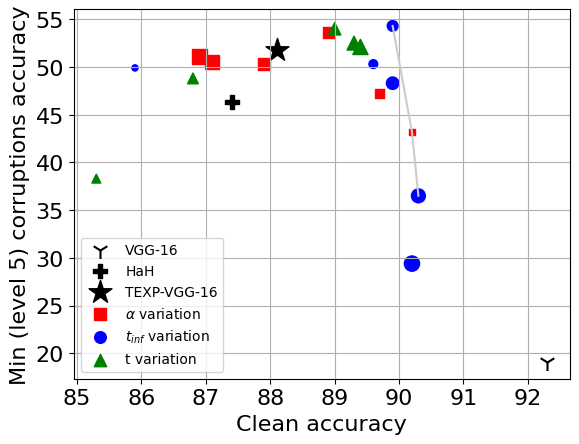}\label{fig:zz_min}}
  \caption{Robustness-accuracy trade-offs obtained by varying the TEXP parameters. Increasing marker sizes represent increasing values of the corresponding TEXP parameter. The red markers with increasing sizes encode $\alpha$ variation for $\alpha \in$ $[0.00001, 0.0001, 0.0005, 0.002, 0.005, 0.01]$. The blue markers with increasing sizes encode $t_{\text{inf}}$ variation, for $t_{\text{inf}}\in $ [$0.5/\sqrt{D}$, $2/\sqrt{D}$, $3/\sqrt{D}$, $4/\sqrt{D}$, $8/\sqrt{D}$,  $16/\sqrt{D}$]. The green markers with increasing sizes encode $t = \beta t_{\text{inf}}$ variation for $\beta \in $ $[1, 5, 15, 25, 50]$. %
  }
  \label{fig:rob_acc_min}
\end{figure}
To study the sensitivity of the robustness-accuracy performance to the TEXP parameters, TEXP objective weight $\alpha$, inference tilt $t_\text{inf}$ and training tilt $t$, we perform the following experiment. While keeping two of the three TEXP parameters fixed to those set in the TEXP-VGG-16 model (where we had $\alpha=0.001$, $t_{\text{inf}}=1/\sqrt{D}=0.192$, $t=\beta t_{\text{inf}}$, where $\beta=10$), we vary the third and train a new TEXP model. We record the clean accuracy and the average corruption robustness (both for all levels and the highest severity level). Fig.~\ref{fig:rob_acc} shows the robustness-accuracy trade-offs of the various TEXP models, where robustness is measured against corruptions of all severity level (Fig.~\ref{fig:z}) and highest severity level (Fig.~\ref{fig:zz}). 
To plot the variation with $\alpha$, we fix $t_{\text{inf}}=0.192$, $\beta=10$, and vary $\alpha$ from $0.00001$ to $0.01$ (with the larger size of the markers in the figures representing larger values of $\alpha$, similarly for other variations). For the variation in $t_{\text{inf}}$, we fix $\alpha=0.001$, $\beta=10$, and vary $t_{\text{inf}}$ from $0.5/\sqrt{D}$ to $16/\sqrt{D}$. For the variation in $t$, we fix $\alpha=0.001$, $t_{\text{inf}}=0.192$, and vary $\beta$ from $1$ to $50$. We also report the trade-offs with respect to the minimum (worst-case) corruption accuracy in Fig.~\ref{fig:rob_acc_min}. Overall, our results show that TEXP is mildly sensitive to the choice of the hyperparameters and all variants of TEXP dominate HaH~\citep{cekic} in the Pareto plane. %

\section{TEXP variant with expanded neural competition }
\label{app:expensive_texp_variant}
In this section, we describe an alternate method of applying the TEXP principles to DNNs. We use the VGG-16 backbone, and change the first layer to a TEXP layer, with a few distinctions in the way tilted softmax, thresholding and TEXP objective is applied.

Similar to the original approach, we implicitly normalize the convolution filter weights to unit $\ell_2$ norm as follows:
\begin{equation}
y_i (l) =  \frac{ (\bx (l))^T \bw_i }{||\bw_i||_2}.
\end{equation}

Next, instead of promoting competition across the $M$ filters through the tilted softmax, we introduce competition among all the activations in the layer output. Let us index the convolution layer outputs $y_i(l)$, across all filters and spacial locations, by $y_{m}, m\in [M']$, where $M'=L\times M$. For the running example in the paper with VGG-16 model and CIFAR-10 inputs, we have $M' = 32\times32\times64$, i.e., the dimension of the layer output post convolution. We then pass these convolution outputs through a tilted softmax, normalizing over the $M'$ scalar layer outputs:%
$$
p_{m} = {\sigma}_{m}(t_\text{inf} \by),
$$ 
where,
$$
{\sigma}_{m}(\bx) = \frac{\exp(x_{m})}{\sum_{j=1}^{M'} \exp(x_j)}.
$$
and $\by = \{y_{m}, m=1,2,\ldots,M'\}$. 

We reindex the post softmax outputs $p_{m}$ by filter $i$ and spatial location $l$ as $p_i(l)$, and use these notations interchangeably.   

Further, in the thresholding block, the thresholds $\tau_i$ are set such that for every image, we permit only a certain fraction of the activations, while nullifying the rest. For instance, we set $\tau_i$ adaptively such that $20\%$ of the outputs are activated for each image, and each filter. This requires sorting of the activations to decide the thresholds, which makes this approach expensive. 

The TEXP objective here is given by
\begin{equation} \label{TEXP_objective_CNN}
    \mathcal{L}_{\text{TEXP}} = \frac{1}{t} \log \left(\frac{1}{M'} \sum_{m = 1}^{M'} \exp({t a_{m}})\right)
\end{equation}
and the \textit{balanced} TEXP objective is 
$$
\mathcal{L}_{\text{TEXP}}^{\text{bal}} = \frac{1}{t} \log \left(\frac{1}{M'} \sum_{m = 1}^{M'} \exp({t (a_{m} - \Bar{a})})\right)
$$
where  $a_{m} = \text{ReLU}(y_{m})$ are the convolution outputs across all filters and spatial locations in the first layer, passed through a ReLU function, and $\Bar{a} = (1/M')\sum_{m = 1}^{M'} a_{m}$ denotes the mean of all the post-ReLU activations in the layer. 

We term this variant the TEXP-v2. We set the tilt parameters as $t_{\text{inf}} = 0.1$, $t=1$ and $\alpha=0.0001$, while other optimization hyperparameters are retained the same. We observe that augmenting this variant with adversarial training while utilizing the balanced TEXP objective resulted in a model that has a better robustness-accuracy trade-off. We present in Table~\ref{tab:var_texp} the test accuracies of this variant contrasted with the matching TEXP model from Section~\ref{sec:texp_dnn}. We clarify that all the models listed in the main paper are trained with the (non-balanced) TEXP objective.

\begin{table*}[t]
\centering 
\resizebox{0.95\textwidth}{!}{
\begin{tabular}{ccccccc}
\toprule
{Model} & {Clean} & {Noise} & {Min/Avg} & {Min/Avg}& {{Autoattack $\ell_2$} } & {{Autoattack $\ell_{\infty}$}}\\
{ } & {} & {$\nu=0.1$ }  & {corruptions} & {severity level: $5$}& {{adv, $\epsilon=0.25$}} & {{adv, $\epsilon=2/255$}}\\ \toprule
VGG-16 & $92.3$ & $24.1$  &$46.9/72.4$ & $19.1/54.6$& { $13.2$ } & {$10.2$}\\
TEXP-VGG-16 & $88.1$ & $75.5$  & $74.1/80.3$ & $51.8/72.5$& {$51.4$} & {$42.0$}\\
TEXP-v2-VGG-16 & $88.3$ & $68.4$  & $69.7/79.6$ & $48.3/71.8$& {$39.4$} & {$27.6$}\\
\midrule
VGG-16 + Adv Tr& $87.8$ & $79.7$  & $51.4/79.1$ & $18.2/70.1$ & {$71.5$ } & {${72.3}$}\\
TEXP-VGG-16 (Balanced) + Adv Tr &$85.8$ & ${81.9}$  & $67.3/80.3$ & $37.2/74.4$& {$69.4$} & {$64.8$}\\
TEXP-v2-VGG-16 (Balanced) + Adv Tr &$89.0$ & $81.1$  & $78.6/84.1$ & $56.9/79.2$& {${71.8}$} & {$63.4$}\\
\bottomrule
\end{tabular}}
\caption{Robustness to corruptions under the variant TEXP-v2 models on CIFAR-10 clean and corruptions datasets. All numbers reported are test accuracies.}
\label{tab:var_texp}
\end{table*}

Although this allows us to explicitly control the sparsity levels at the end of first layer and promotes competition among the layer outputs directly, we recommend the TEXP approach described in Section~\ref{sec:texp_dnn} as it aligns more closely to learning matched filters and avoids sorting of activations during training and inference to set the thresholds. For completeness, the average training (inference) time for 1 epoch on the TEXP-v2-VGG-16 model is $31.1$ sec ($3.19$ sec) on an NVIDIA 1080 Ti GPU.

\section{Sparsity of activations after the first network layer}
\label{app:texp_dnn_details}
\begin{figure}[h]
  \centering 
  \subfigure[Sparsity in $C\times H\times W$ dimensions]{\includegraphics[width=0.331\columnwidth]{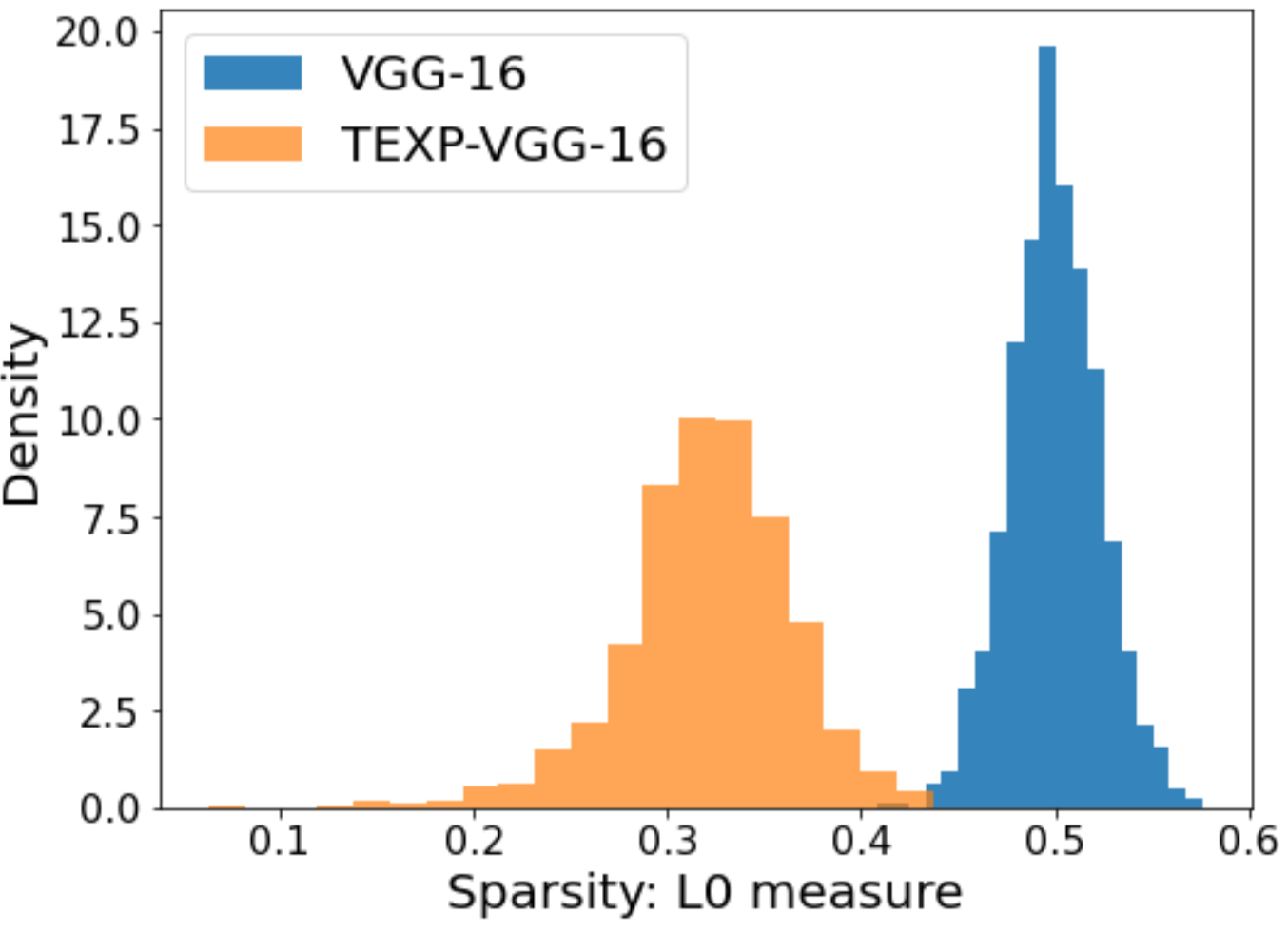}\label{fig:lay_out}}
  \hfill
  \subfigure[Sparsity in channel/filter dimension]{\includegraphics[width=0.318\columnwidth]{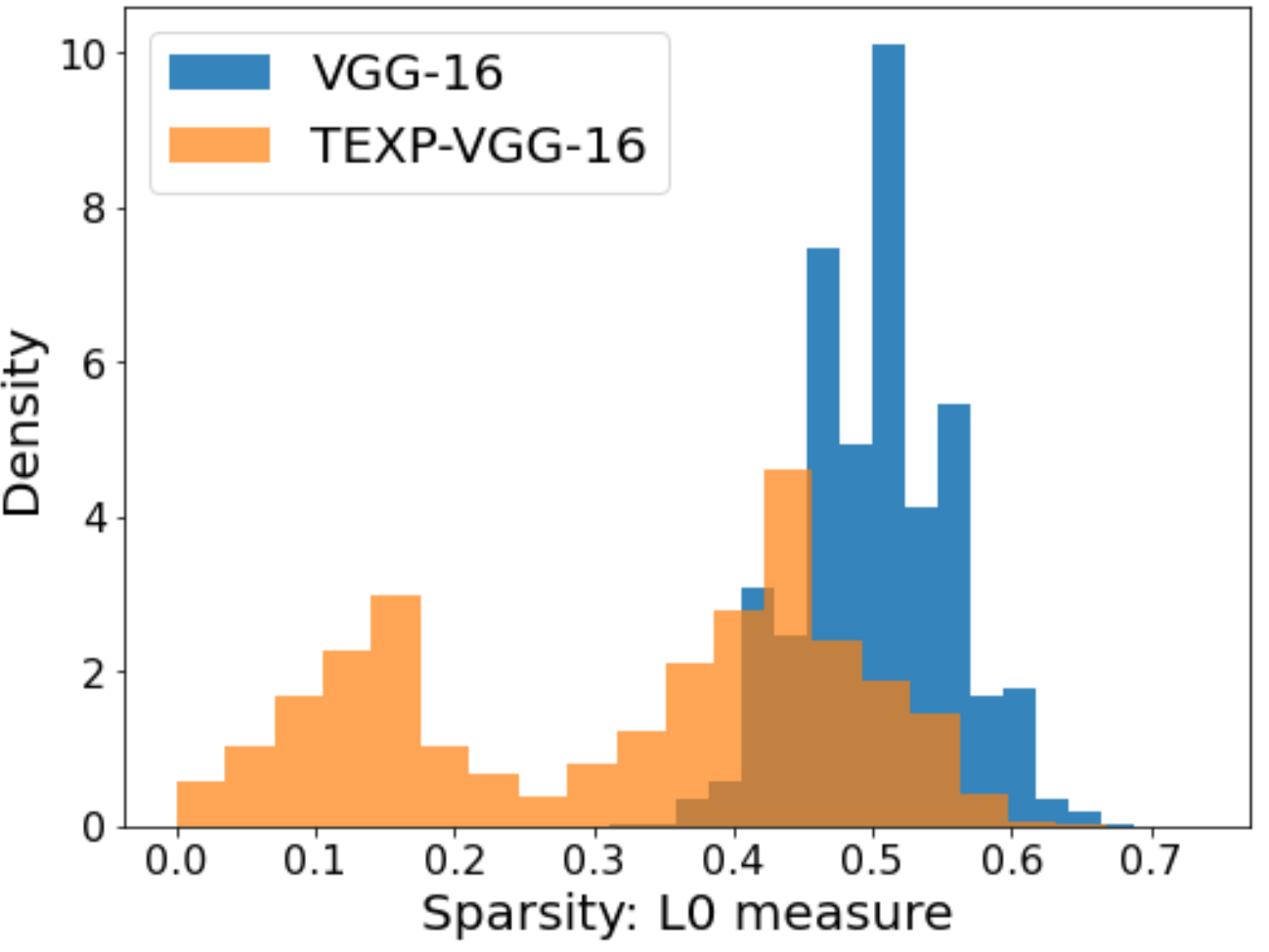}\label{fig:bb}}
  \hfill
  \subfigure[Sparsity in spatial dimensions]{\includegraphics[width=0.315\columnwidth]{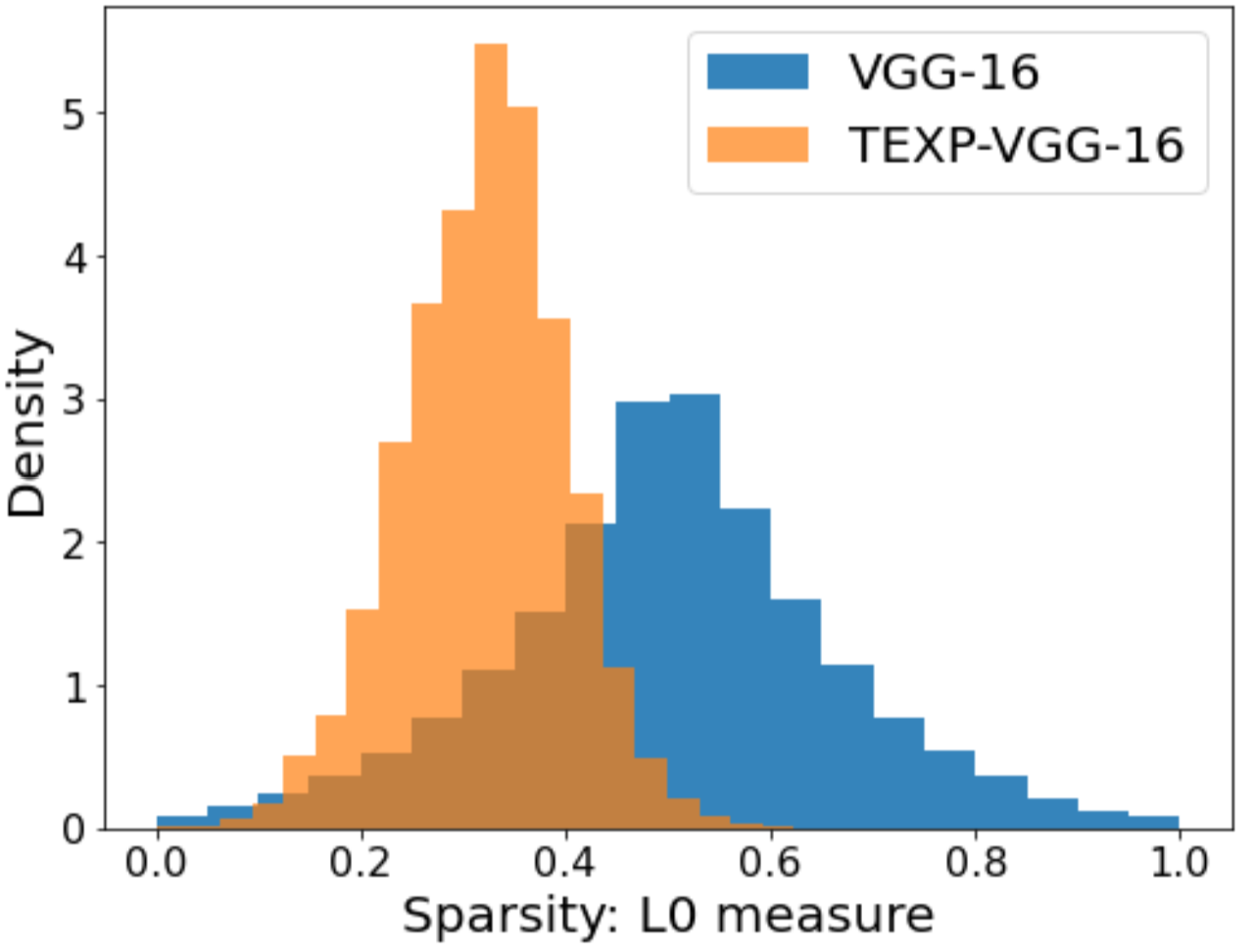}\label{fig:cc}}
  \caption{Histograms of the sparsity levels in the first layer output (lower values indicate more sparse).}
  \label{fig:spar_dnn}
\end{figure}

In this section, we show the statistics of sparsity of the first layer outputs for both VGG-16 and TEXP-VGG-16 models. Recall that in the baseline VGG-16, the ordering is as follows: convolution, followed by ReLU, and then batch-norm. Since batch-norm renders the output completely non-sparse, to draw a fair comparison with TEXP, we obtain sparsity statistics post the ReLU, for VGG-16. For TEXP-VGG-16, we measure the sparsity statistics on the output of the thresholding block in the TEXP layer. We wish to obtain insights on the following questions through experiments: (i) how sparse is the layer output? (ii) for each spatial location, how does neuronal competition give rise to sparsity in the channel/filter dimension? (iii) for each filter, how sparse is the output across the spatial locations? To evaluate these, we take a batch of $1000$ CIFAR-10 test images, and compute an $L_0$ notion of sparsity on the first layer output as the number of non-zero entries (measured by counting activations larger than some small epsilon) normalized by the number of scalar dimensions. For instance, to answer the first question, given an image, we count the number of non-zeros over the entire layer output and normalize by $C\times H\times W$. For the second question, for every spatial location, we count the non-zeros along channel dimension, and normalize by the number of channels $C$. For the third, for every filter, we count the non-zeros in spatial dimensions and normalize by $H\times W$. We then plot the histograms of the sparsity levels obtained for the three cases in Fig.~\ref{fig:spar_dnn}. We observe that TEXP leads to sparse layer outputs, as expected, due to the neuronal (and spatial) competition from TEXP inference and training.

\section{Statistical significance of the results}
\label{app:stat_sig}
The results in Table~\ref{tab:performance} on the performance of all the VGG based models on CIFAR-10 dataset are reported after averaging across five runs. The numbers reported are the sample mean values of the accuracies across the five runs $\pm$ standard error on the mean. The standard error is given by $\sigma/\sqrt{N}$, where $\sigma$ is the sample standard deviation and $N$ is the number of runs. %

\section{Ablation study}\label{app:abl_study}
\edits{In this section, we contrast the contributions of the TEXP objective and components of the inference paths in comparison to the baseline model. We show in Table~\ref{tab:ablation} four VGG-$16$ based models: (i) Baseline VGG; (ii) VGG with the adpative thresholding block replacing the ReLU in the first layer (named VGG-16 + Thresholding), which is similar to a TEXP inference layer albeit with retaining the batchnorm; (ii) TEXP-VGG-16-Inference-only, which introduces the tilted softmax in place of the batchnorm (and with $\alpha$ = 0, i.e. no TEXP objective); (iii) typical TEXP-VGG-16, which introduces the TEXP objective with $\alpha>0$ over the inference-only structure. We can observe that addition of each of the components of the TEXP approach improves the robustness of the model.

Although there have been works in literature that approach robustness through stochastic pruning techniques, we observe that activation pruning alone through the thresholding block (which performs deterministic activation pruning), has a limited impact on robustness. However, when the thresholding and tilted softmax blocks are combined with the TEXP objective, our approach aligns filters to match the incoming patterns and shapes the activations, ensuring that there is some fraction of strong activations that survive the thresholding.

}

\begin{table*}[h]
\centering 
\resizebox{\textwidth}{!}{
\begin{tabular}{cccccccc}
\toprule
{Model} & {Clean} & {Noise} & {Min/Avg} & {Min/Avg}& {{Autoattack $\ell_1$} } & {{Autoattack $\ell_2$} } & {{Autoattack $\ell_{\infty}$}}\\
{ } & {} & {$\nu=0.1$ }  & {corruptions} & {severity level: $5$}& {{adv, $\epsilon=3$}}& {{adv, $\epsilon=0.25$}} & {adv, $\epsilon=2/255$}\\ \toprule
VGG-16 & $92.3$ & $24.1$  &$46.9/72.4$ & $19.1/54.6$&$10.1$ &{$13.2$} & {$10.2$}\\
VGG-16 + Thresholding& $91.6$ & $39.1$  &$53.1/73.8$ & $18.4/55.7$& $15.1$&{ $20.5$ } & {$16.1$}\\
TEXP-VGG-16-Inference-only & $90.0$ & $60.8$  & $63.1/78.7$ &$44.0/68.3$ &$37.0$&$36.3$&$24.5$ \\
TEXP-VGG-16 & $88.1$ & $75.5$  & $74.1/80.3$ & $51.8/72.5$& $47.2$&{$51.4$} & {$42.0$}\\
\bottomrule
\end{tabular}}
\caption{Ablation study of TEXP components on CIFAR-10 clean and corruption datasets. All numbers reported are test accuracies for a single run.}
\label{tab:ablation}
\end{table*}

\section{Computation time}
\label{app:compute}
All the experiments on the CIFAR-10 dataset were mainly performed using $4$ NVIDIA GeForce 1080 Ti GPUs with 12GB memory. For experiments on the CIFAR-100 and ImageNet-1K datasets, we relied on an NVIDIA A100 GPU with 80GB memory. The training and inference times for the baseline VGG-16, HaH and TEXP-VGG-16 models are contrasted in Table~\ref{tab:runtimes}. To train HaH models, we use the repository by~\cite{cekic}, and utilize the set of hyperparameters outlined in the paper.

\begin{table*}[h]
\centering
\resizebox{0.6\textwidth}{!}{
\begin{tabular}{ccccc}
\toprule
{  GPU $\rightarrow$} & \multicolumn{2}{c}{1080Ti} & \multicolumn{2}{c}{A100}\\
{ Models $\downarrow$} & Training &Inference & Training&Inference\\
\toprule
VGG-16 & 18.9$\pm$0.27 & 2.79$\pm$0.02 & 5.9$\pm$0.08 & 1.13$\pm$0.006 \\
HaH~\citep{cekic} & 107.3$\pm$0.38 & 4.00$\pm$0.04 & 34.6$\pm$0.06 & 1.41$\pm$0.006 \\
TEXP-VGG-16 & 27.5$\pm$0.18 & 2.84$\pm$0.01 & 8.23$\pm$0.15 & 1.12$\pm$0.009 \\
\bottomrule
\end{tabular}}
\caption{Average training (for 1 epoch) and inference times in seconds. The numbers reported are average time $\pm$ standard error in mean, where standard error = sample standard deviation/$\sqrt{N}$.}
\label{tab:runtimes}
\end{table*}

\end{document}